\documentclass[10pt,twocolumn,letterpaper]{article}

\usepackage[pagenumbers]{cvpr} 

\usepackage[T1]{fontenc}
\usepackage{multirow}
\usepackage{booktabs}
\usepackage{microtype}
\usepackage{placeins}
\definecolor{cvprblue}{rgb}{0.21,0.49,0.74}
\usepackage[pagebackref,breaklinks,colorlinks,allcolors=cvprblue]{hyperref}
\def\confName{CVPR}
\def\confYear{2027}

\definecolor{cvprblue}{rgb}{0.21,0.49,0.74}
\usepackage[pagebackref,breaklinks,colorlinks,allcolors=cvprblue]{hyperref}

\def\confName{CVPR}
\def\confYear{2027}

\title{AeroLLE: Nighttime Aerial Image Enhancement with Generated Supervision}

\author{Wei Lu$^{1}$, Hongyuan Liu$^{1}$, Liuqian Wang$^{2}$, Qi Ming$^{3}$, Xue Yang$^{4}$, Si-Bao Chen$^{1}$$^\ast$\\ \\
	$^{1}$School of Computer Science and Technology, Anhui University, Hefei 230601, China \\ $^{2}$School of Cyber Science and Engineering, Zhengzhou University, Zhengzhou 450001, China \\ $^{3}$College of Computer Science, Beijing University of Technology, Beijing 100124, China \\ $^{4}$School of Automation and Intelligent Sensing, Shanghai Jiao Tong University, Shanghai 200240, China\\
	{\tt\small luwei@ahu.edu.cn, 1287842367@qq.com, sbchen@ahu.edu.cn}
}

\begin{document}
\maketitle

\begin{abstract}
Nighttime aerial image enhancement is constrained by the scarcity of aligned normal-light targets. Acquiring such targets from moving UAVs is difficult, and simple darkening of daytime images underrepresents the sensor noise, localized lighting, and uneven exposure of real nighttime captures. Generating normal-light references from real nighttime observations offers a practical way to fill this supervision gap. AeroNight pairs 1,200 real nighttime aerial RGB images with generated references selected through manual screening, preserving real imaging degradation in the inputs. This setting motivates AeroLLE, a lightweight two-stage framework whose context-aware Base Enhancer combines local processing with scene-level frequency information to recover visibility. Spatially Adaptive Exposure-Color Calibration (SAECC) then refines the base prediction through bounded low-resolution RGB gain and bias fields. AeroLLE achieves the highest PSNR and SSIM among the evaluated methods on both held-out test sets, reaching 17.62/17.89 dB PSNR and 0.5190/0.4964 SSIM on Test 1/Test 2, respectively. The model uses 255.24K parameters and 2.47G FLOPs, reducing FLOPs by 85.4\% relative to Multinex and improving PSNR by 1.56 dB on Test 2. AeroNight provides the first benchmark for pseudo-supervised nighttime aerial enhancement, and AeroLLE shows that generated references can support compact enhancement models.
\end{abstract}

\section{Introduction}
\label{sec:introduction}

Nighttime aerial imaging supports urban monitoring and infrastructure inspection under sparse and uneven illumination. A single frame often contains barely visible roads, roofs, and vegetation beside saturated streetlights, with distant objects and thin boundaries occupying only a few pixels. Enhancement aims to reveal poorly illuminated regions while controlling highlight spread, color distortion, and noise amplification. Recent advances in color representations, degradation modeling, and illumination priors have improved low-light enhancement models \cite{HVI,DarkIR,URWKV,Multinex}. For supervised nighttime aerial enhancement, obtaining aligned normal-light targets for the scenes captured by UAVs at night remains a persistent data acquisition challenge.

\begin{figure}[t]
	\centering
	\includegraphics[width=\linewidth]{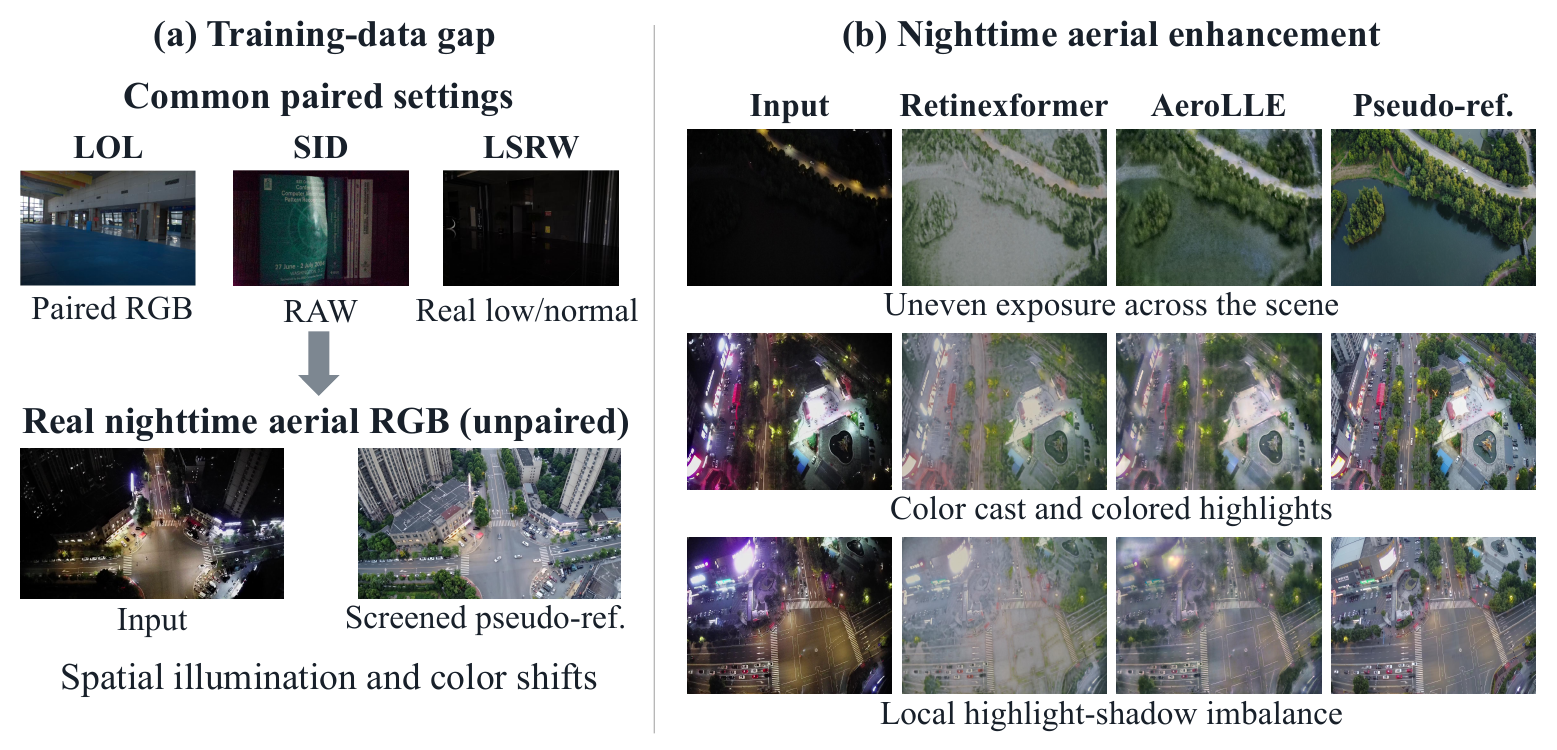} \vspace{-7mm}
	\caption{Data construction and representative enhancement results. (a) LOL \cite{LOL}, SID \cite{SID}, and LSRW \cite{LSRW} provide paired RGB, short/long-exposure RAW, and real low/normal-light captures, respectively. AeroNight pairs real nighttime aerial RGB inputs with screened generated normal-light references. (b) Examples show uneven exposure, colored illumination, and local light imbalance.} \vspace{-6mm}
	\label{fig:motivation}
\end{figure}

Paired datasets such as LOL, SID, and LSRW have supported progress in low-light enhancement \cite{LOL,SID,LSRW}. Collecting comparable supervision from moving UAVs involves matching viewpoints and scene content across captures despite changes in traffic, local illumination, and environmental conditions. Synthetic darkening of processed normal-light photographs provides registered pairs with limited coverage of the sensor noise, localized light sources, and uneven exposure present in real nighttime captures. Denoising, tone mapping, and compression can further weaken the degradation cues retained in the source photographs. These constraints highlight the importance of constructing training pairs that reflect real aerial capture conditions.

A real nighttime observation already contains the degradation that the enhancer is intended to handle, which motivates generating a normal-light reference from that observation. Generation supplies the missing target while the captured input retains the noise, light-source effects, and exposure distribution encountered during aerial imaging. Reference generation and screening thus provide a practical route to building supervised training data from real nighttime flights. Figure \ref{fig:motivation} illustrates this data setting and representative enhancement challenges.

AeroNight establishes this setting with 1,200 pairs at 2736$\times$1536 resolution, each combining a real nighttime UAV RGB image with a generated normal-light reference selected through manual screening. The scenes cover roads, residential areas, buildings, bridges, and vegetation across different aerial viewpoints. Screening rejects candidates with visible content changes, geometric discrepancies, or implausible illumination and color. Two held-out test sets evaluate new images from locations represented during development and images from previously unseen locations. This benchmark provides a common basis for studying real aerial low-light enhancement, generated supervision, and generalization across locations.

Learning from these pairs calls for scene-wide illumination context together with local image processing. Large dark regions and compact light sources can require different corrections within the same frame. AeroLLE addresses these regional differences through a lightweight two-stage framework. Its context-aware Base Enhancer combines a full-resolution convolutional path with a scene-thumbnail path, keeping scene-context computation on a small grid. Cross-Spectral Context (CSC) performs band-specific mixing of latent channels within four spatial-frequency bands. The resulting context guides local feature modulation and spatial exposure correction, giving the full-resolution path access to the illumination pattern of the entire scene.

The remaining exposure and color errors in the base prediction also vary spatially. Spatially Adaptive Exposure-Color Calibration (SAECC) represents their correction through bounded low-resolution RGB gain and bias fields. With the selected Base Enhancer frozen, a coarse affine teacher derives field-level targets from the pseudo-reference to guide calibration learning. Bilinear upsampling extends the predicted fields to the original image resolution, giving the second stage a smooth representation of exposure and color adjustments. Reference generation takes place during dataset construction, and inference uses a single nighttime RGB image.

On two held-out test sets, AeroLLE achieves the highest PSNR and SSIM among the evaluated methods. Its PSNR values of 17.62 and 17.89~dB exceed CIDNet by 1.21 and
0.64~dB, respectively. The complete inference pipeline uses 255.24K parameters and
2.47G FLOPs at 512$\times$512 resolution, reducing FLOPs by 85.4\% relative to Multinex. On an RTX 5090, it processes full-resolution 2736$\times$1536 images at 62.12 FPS. Qualitative results show improved dark-region visibility and more balanced exposure and color across mixed nighttime illumination. These results demonstrate the potential of generated references for learning compact enhancement models from real aerial observations.

The main contributions are summarized as follows.
\begin{itemize}
	\item AeroNight pairs 1,200 real nighttime aerial RGB images with generated normal-light references selected through manual screening. It preserves real imaging degradation in the inputs and establishes the first benchmark for nighttime aerial RGB image enhancement with generated-reference supervision and evaluation across locations.
	
	\item AeroLLE couples context-aware visibility recovery with spatial exposure-color calibration in a lightweight two-stage framework. CSC supplies scene-level context through band-specific latent-channel interactions, and SAECC refines the base prediction through bounded low-resolution RGB affine fields guided by a coarse teacher.
	
	\item Experiments on seen- and unseen-location test sets show that screened generated references support enhancement learning, with AeroLLE achieving the highest PSNR and SSIM among the evaluated methods on both sets. These findings motivate reference generation as a practical route to expanding supervision for aerial enhancement where aligned normal-light targets are difficult to acquire.
\end{itemize}

\section{Related Work}
\label{sec:related_work}
Low-light datasets provide paired low/normal-light RGB
images~\cite{LOL,LSRW}, short/long-exposure RAW
captures~\cite{SID}, and multi-exposure observations~\cite{SICE}.
ExDark~\cite{ExDark} collects real low-light images for recognition.
Specialized datasets extend coverage to underwater
stereo~\cite{SeaLowLight}, thermal--visible pairs~\cite{ThermalLL},
and event--image observations~\cite{EventLL,EICLIE,BiEvLight}.
For UAV imaging, changes in viewpoint, traffic, and illumination
make aligned normal-light targets difficult to acquire.
AeroNight generates and screens normal-light references for
captured nighttime aerial images, providing pseudo-paired
supervision that retains real low-light inputs.

Illumination and degradation modeling guide many enhancement methods.
Retinex-based approaches incorporate illumination modeling into
restoration~\cite{Retinexformer,IniRetinex,KinDpp}, and
RUAS~\cite{RUAS} combines Retinex unrolling with architecture search.
CIDNet~\cite{HVI} introduces an HVI color space for processing
chromatic information and intensity.
Reference-free methods formulate enhancement through learned
curves~\cite{ZeroDCE}, self-calibrated illumination~\cite{SCI,SCIpp},
or physical priors~\cite{QuadPrior}.
Specularity factorization~\cite{Specularity} models illumination
and reflectance effects.
SNR-Aware~\cite{SNRAware} accounts for spatially varying noise,
and DarkIR~\cite{DarkIR} jointly addresses low-light degradation
and blur.

\begin{figure*}[t]
	\centering
	\includegraphics[width=\linewidth]{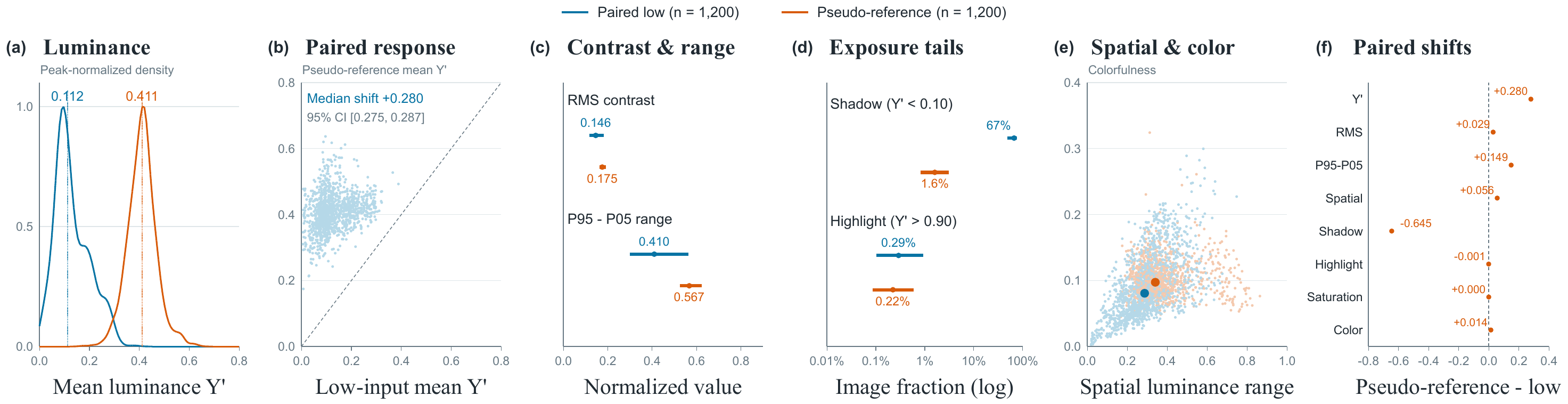} \vspace{-6mm}
	\caption{Exposure and contrast statistics of AeroNight. Statistics cover all 1,200 pairs. (a) Peak-normalized kernel density estimates of mean luminance. (b) Paired input and pseudo-reference mean luminance. (c) RMS contrast and P95-P05 luminance range. (d) Shadow-pixel (\(Y^\prime<0.10\)) and highlight-pixel (\(Y^\prime>0.90\)) fractions. (e) Spatial luminance range and colorfulness. (f) Median reference-minus-input changes with 95\% bootstrap confidence intervals from 10,000 image-pair resamples. Bars in (c,d) indicate interquartile ranges. Images are area-downsampled to a maximum side of 512 pixels. Luminance is \(Y^\prime=0.299R+0.587G+0.114B\). 
		} \vspace{-3mm}
	\label{fig:statistics}
\end{figure*}

High-resolution enhancement requires efficient processing and access to scene context. MobileIE~\cite{MobileIE} uses a compact re-parameterizable ConvNet, SPJFNet~\cite{SPJFNet} introduces joint frequency enhancement, and Multinex~\cite{Multinex} combines lightweight processing with multiple Retinex priors. URWKV~\cite{URWKV} captures long-range dependencies through multi-state representations. Implicit representations~\cite{ImplicitLL} and semantic guidance~\cite{SemanticLL} provide complementary image and task information. AeroLLE uses band-specific channel interactions on a scene thumbnail to guide full-resolution local corrections with scene-wide illumination context.

Generative enhancement learns low-to-normal mappings through adversarial training \cite{EnlightenGAN}, normalizing flows \cite{LLFlow}, and diffusion models \cite{ExposureDiffusion,ReDDiT,BiBridge,MR,ZeroIDIR}. Generative perceptual priors guide restoration through learned image distributions \cite{GenerativeLL}. Foundation models also provide guidance during enhancement training, with FoCo \cite{FoCo} using collaborating vision foundation models to improve semantic features and downstream performance. AeroNight uses image generation during dataset construction to obtain screened normal-light references for real nighttime aerial inputs. These pairs supervise the compact AeroLLE enhancer, whose second stage represents exposure and color refinement through bounded low-resolution RGB affine fields.

\section{AeroNight Dataset}
\label{sec:dataset}
AeroNight pairs 1,200 real nighttime UAV RGB images with generated normal-light references selected through manual screening at 2736$\times$1536 resolution.
Scenes cover urban roads, residential areas, buildings, bridges and vegetation from varied aerial viewpoints. The captured inputs contain uneven exposure, mixed artificial lighting, sensor noise, and compression artifacts
recorded during nighttime flight. Figure~\ref{fig:statistics} summarizes the exposure and contrast distributions of the inputs and pseudo-references.

The median per-image mean luminance increases from 0.112
in the nighttime inputs to 0.411 in the references.
The median paired increase is \(0.280\), with a 95\%
image-pair bootstrap confidence interval of \([0.275,0.287]\).
Median RMS contrast rises from 0.146 to 0.175 and the median
P95--P05 luminance range increases from 0.410 to 0.567.
The median shadow-pixel fraction (\(Y^\prime<0.10\))
falls from 67\% to 1.6\%.
The median highlight-pixel fraction (\(Y^\prime>0.90\))
remains small, changing from 0.29\% to 0.22\%.

\begin{figure}[t]
	\centering
	\includegraphics[width=\linewidth]{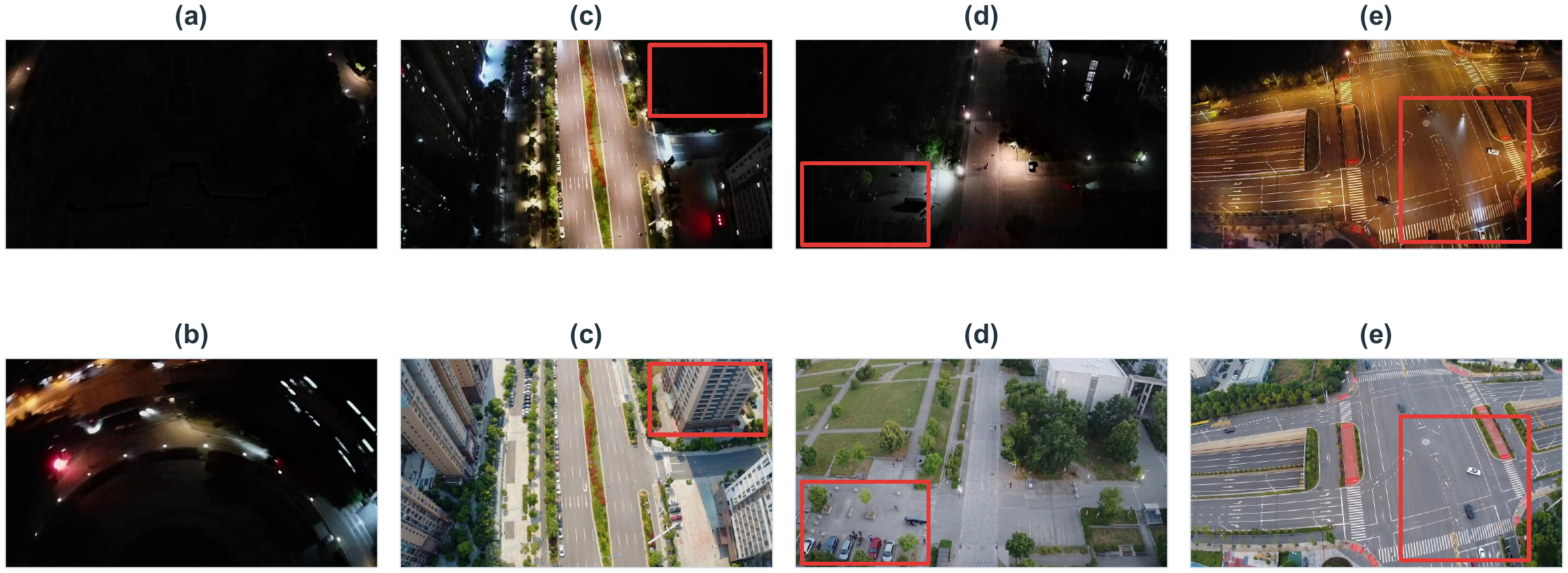} \vspace{-7mm}
	\caption{Rejected cases in AeroNight screening. (a,b) Inputs with insufficient visibility or severe blur. (c-e) Generated candidates with synthesized building details, changes in a parking area, and altered intersection markings. In (c-e), nighttime inputs appear above generated candidates. Red boxes mark corresponding regions with content changes.} \vspace{-6mm}
	\label{fig:screening}
\end{figure}

These pseudo-references are constructed through image generation and manual screening. Gemini 3.1 Pro generates more than 1,600 normal-light candidates from the nighttime inputs, of which 1,200 are retained. Inputs with insufficient visibility or severe blur are excluded. Reference screening rejects candidates with added or removed objects, rewritten text or markings, or visible geometric misalignment. Candidates with implausible illumination or color are also excluded. The screened candidates form the pseudo-reference targets used for training and evaluation. Figure \ref{fig:screening} shows representative excluded inputs and generated candidates.

\begin{figure*}[t]
	\centering
	\includegraphics[width=\linewidth]{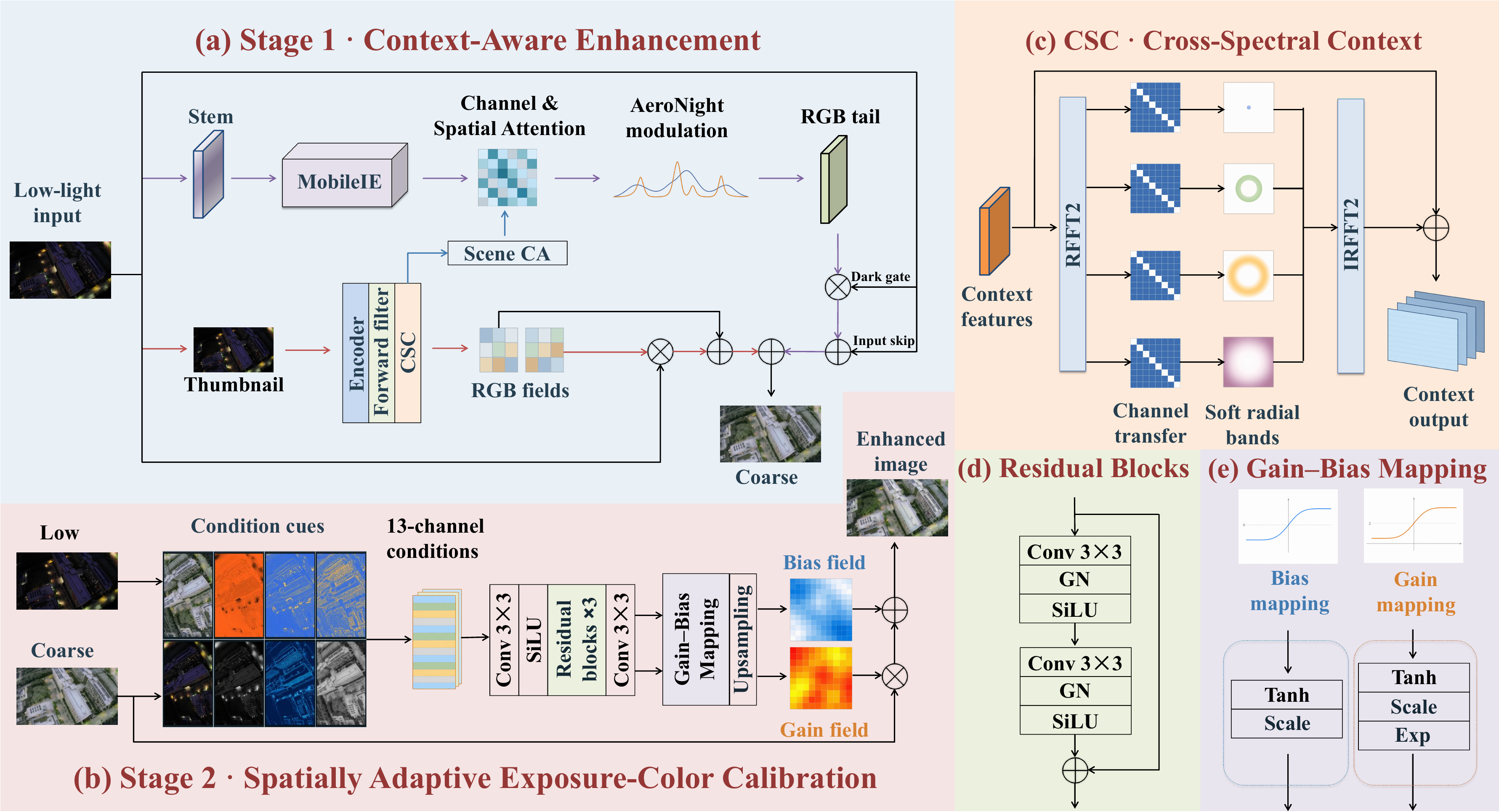}
	\caption{Overview of AeroLLE. (a) The Base Enhancer combines full-resolution local features with thumbnail scene context to estimate \(B_0\). (b) SAECC predicts bounded low-resolution RGB gain and bias fields from the input and base estimate. (c) CSC mixes channels within four fixed soft radial frequency bands. (d,e) Calibration residual blocks and bounded gain-bias mappings.}
	\label{fig:architecture}
\end{figure*}

The dataset is divided into 900 training pairs,
100 validation pairs for checkpoint selection,
and two 100-pair held-out test sets.
Test~1 contains held-out images from locations represented
in the training and validation pool.
Test~2 contains images from previously unseen locations.
This design tests whether enhancement learned from generated
references transfers across locations.

\section{AeroLLE Method}
\label{sec:method}

\subsection{Overview}

Wide aerial views bring roads, roofs, and vegetation illuminated
by different light sources into the same frame.
Enhancement therefore requires scene-wide illumination context
as well as local image evidence.
AeroLLE first recovers visibility with a context-aware Base Enhancer
and then refines the remaining exposure and color imbalance through
spatial calibration.

Given a nighttime aerial RGB image
\(x\in[0,1]^{H\times W\times3}\), let \(F_\theta\) denote the
Base Enhancer and \(H_\phi\) the calibration network including
field upsampling.
With \(\Pi\) denoting element-wise clipping to \([0,1]\),
the two-stage mapping is
\begin{equation}
	\begin{aligned}
		B_0 &= Q\bigl(\Pi(F_\theta(x))\bigr), \\
		(G,A) &= H_\phi(x,B_0), \\
		B_1 &= \Pi(B_0\odot G+A).
	\end{aligned}
	\label{eq:pipeline}
\end{equation}
Here \(\odot\) denotes element-wise multiplication and
\(Q(v)=\lfloor255v+0.5\rfloor/255\) is the 8-bit quantization
between stages.
The stages are trained sequentially.
The selected Base Enhancer is frozen during calibration training,
giving the second stage a fixed image estimate \(B_0\) around
which to learn spatial exposure and color adjustments.
We denote the screened training pseudo-reference by \(y_p\).
Inference applies both stages to a single nighttime RGB image.

\subsection{Context-Aware Base Enhancer}

Small objects and thin boundaries in aerial images call for
processing on the original pixel grid.
Illumination differences across distant regions also require
a view of the entire scene.
The Base Enhancer couples a compact full-resolution path with
a thumbnail context path to meet these requirements.

The full-resolution path adapts the MobileIE
backbone~\cite{MobileIE} with 12-channel features and
re-parameterizable convolutions.
Scene-channel attention and spatial attention modulate the
local features.
Input luminance and local boundary cues further guide RGB
residual prediction.
A luminance-dependent gate scales the residual, and an input
skip connection carries the nighttime image to the output.
The training branches of the re-parameterizable convolutions
are fused for deployment.

The context path receives an aspect-preserving thumbnail
with a maximum side length of 384 pixels.
A compact encoder produces 24-channel features at approximately
one-quarter of the thumbnail resolution.
A residual latent filter aggregates spatial information with
normalized \(5\times5\) kernels.
CSC then performs frequency-dependent channel mixing on these
features.
The resulting representation supplies a scene-channel attention
(Scene CA) vector and six spatial correction channels.
This path provides scene-wide guidance on a small spatial grid.

\subsection{Cross-Spectral Context}

Nighttime aerial images contain illumination variations over
multiple spatial scales, from broad dark surfaces to localized
transitions around artificial light sources.
CSC allows latent-channel interactions to vary with spatial
frequency, giving the context representation different channel
transforms for these scales.

For a context feature \(F\) with \(C=24\) channels, let
\(\widehat F=\operatorname{RFFT2}(F)\), where RFFT2 is the
two-dimensional real fast Fourier transform and IRFFT2
is its inverse.
Four fixed, nonnegative soft radial masks \(m_b\) sum to one
at each frequency.
The band-specific channel transform is
\begin{equation}
	\begin{aligned}
		W_b =
		\frac{0.12}{\sqrt C}
		\tanh(T_b)\odot(J-I), \hspace{1mm}
		b \in\{1,2,3,4\}.
	\end{aligned}
	\label{eq:transfer}
\end{equation}
The transformed features are combined through a residual update
\begin{equation}
	F_x =
	F+\operatorname{IRFFT2}\!
	\left(
	\sum_{b=1}^{4}
	m_b\odot(W_b\widehat F)
	\right).
	\label{eq:csc}
\end{equation}
Here \(T_b\) is a learnable channel-mixing matrix,
\(J\) is the all-ones matrix, and \(I\) is the identity matrix.
Each \(W_b\) acts on the channel vector at each frequency,
with \(m_b\) broadcast over channels.
The mask \(J-I\) removes diagonal terms and assigns the learned
update to cross-channel mixing.
Zero initialization of \(T_b\) makes CSC begin as the identity.
The four bands contain 2,208 active off-diagonal coefficients
and operate on the thumbnail features.

The CSC features produce scene-channel attention and spatial
correction maps.
Attention modulates local features, and the resized maps
adjust exposure and color.
Let \(r_\theta(x)\) denote the local RGB correction,
\(Y(x)\) the input luminance, and \(M^g\) and \(M^a\)
the two resized RGB correction maps.
The unclipped base prediction is
\begin{equation}
	\begin{aligned}
		R ={}& x+e+
		2\tanh\!\bigl(
		r_\theta(x) \\
		&\qquad\odot
		[0.75+1.25(1-Y(x))]
		\bigr),
	\end{aligned}
	\label{eq:local}
\end{equation}
\begin{equation}
	\begin{aligned}
		F_\theta(x) ={}
		R+0.5\tanh(M^g)\odot x 
		+0.5\tanh(M^a).
	\end{aligned}
	\label{eq:base}
\end{equation}
The learned scalar \(e\) controls the global exposure offset. The luminance-dependent factor modulates the local residual according to input brightness. \(M^g\) and \(M^a\) supply scene-conditioned
multiplicative and additive corrections to \(x\).

\subsection{SAECC }

The exposure and color adjustments required by the base estimate
vary across an aerial scene.
Dark vegetation and an illuminated road can require different
corrections even when both are already visible.
SAECC represents this refinement through spatial RGB gain and
bias fields, allowing exposure and channel balance to be
optimized around the fixed base prediction.

The fields are predicted on a coarse grid of size
\(h=\lceil H/32\rceil\) and \(w=\lceil W/32\rceil\). A 13-channel conditioning tensor combines the area-downsampled input
and base estimate with illumination, darkness, gradient,
detail, and relative-illumination cues.
It contains six RGB channels from \(x\) and \(B_0\),
two illumination channels, one dark map, one edge map,
two detail maps, and one exposure-ratio channel.
A width-64 convolutional head with three residual blocks
predicts six raw channels, split into \(z^g\) and \(z^a\).
The residual blocks use group normalization (GN)
and the sigmoid linear unit (SiLU).

The gain and bias fields are parameterized as
\begin{equation}
	\begin{aligned}
		g &= \exp(0.28\tanh(z^g)), \\
		a &= 0.08\tanh(z^a), \\
		G &= U(g), \qquad A=U(a),
	\end{aligned}
	\label{eq:fields}
\end{equation}
where \(U\) denotes bilinear upsampling to the base-image
resolution.
The gain lies within
\([\exp(-0.28),\exp(0.28)]\) and the bias within
\([-0.08,0.08]\).
Bilinear upsampling preserves these amplitude bounds.
The output convolution is initialized to zero, giving
\(g=1\) and \(a=0\) at the start of training.
The coarse grid and field-smoothness penalty encourage
gradual spatial changes, and independent RGB fields allow
exposure and color to be adjusted jointly.

\subsection{Two-Stage Objectives}

\textbf{Base-enhancement objective.}
Stage 1 learns from the screened pseudo-reference using
the raw prediction \(p=F_\theta(x)\) before clipping
or quantization.
With \(d=p-y_p\), the objective is
\begin{equation}
	\begin{aligned}
		\mathcal L_1 ={}&
		\operatorname{mean}(d^2)
		+0.10\operatorname{mean}(|d|) +0.05\mathcal L_{\mathrm{grad}} \\
		&	+0.01\mathcal L_{\mathrm{mean}}+0.01\mathcal L_{\mathrm{range}}
		+0.10\mathcal L_{\mathrm{under}}.
	\end{aligned}
	\label{eq:stage1}
\end{equation}
The gradient term \(\mathcal L_{\mathrm{grad}}\)
penalizes luminance-gradient differences and
\(\mathcal L_{\mathrm{mean}}\) aligns RGB channel means.
The range term \(\mathcal L_{\mathrm{range}}\)
penalizes predictions outside \([0,1]\).
The underexposure term \(\mathcal L_{\mathrm{under}}\)
weights reference midtones and penalizes a luminance deficit
beyond \(0.02\).

\textbf{Calibration objective.}
The pseudo-reference also defines explicit targets for
spatial calibration.
Let \(D_{32}\) denote area downsampling to the SAECC grid,
with \(\bar B_0=D_{32}(B_0)\) and
\(\bar y_p=D_{32}(y_p)\).
A coarse affine teacher converts their differences into
gain and bias targets
\begin{equation}
	\begin{aligned}
		g^* &=
		\operatorname{clip}\!\left[
		\frac{\bar y_p}
		{\max(\bar B_0,10^{-3})},
		0.7,1.6
		\right], \\
		a^* &=
		\operatorname{clip}\!\left(
		\bar y_p-g^*\odot\bar B_0,
		-0.08,0.08
		\right), \\
		B^* &=
		\Pi\bigl(
		B_0\odot U(g^*)+U(a^*)
		\bigr).
	\end{aligned}
	\label{eq:teacher}
\end{equation}
Constructing these targets on the field grid expresses the reference appearance as a coarse photometric adjustment. SAECC is trained to approximate the teacher fields and image target \(B^*\) using the bounded parameterization.
\begin{equation}
	\begin{aligned}
		\mathcal L_2 ={}&
		\mathcal L_{\mathrm{img}}
		+1.2\mathcal L_{\mathrm{gain}}
		+0.5\mathcal L_{\mathrm{bias}} \\
		&+0.004\mathcal L_{\mathrm{TV}}
		+\rho(k)\mathcal L_{\mathrm{direct}}.
	\end{aligned}
	\label{eq:stage2}
\end{equation}
Here \(\mathcal L_{\mathrm{img}}\) is the L1 distance
between \(B_1\) and \(B^*\).
The gain and bias terms are smooth-L1 losses comparing
\((g,a)\) with \((g^*,a^*)\).
The smoothness term \(\mathcal L_{\mathrm{TV}}\)
measures the mean absolute horizontal and vertical variation
of the coarse fields.
The direct-reference term \(\mathcal L_{\mathrm{direct}}\)
combines image MSE, L1, and dark-region L1 losses against
\(y_p\) with weights \(0.04\), \(0.03\), and \(0.10\).

Calibration training first follows the coarse affine teacher
and then gradually incorporates direct pseudo-reference
supervision.
For calibration epoch \(k\), we set \(\rho(k)=0\)
during the first 24 epochs and increase it linearly to one
over the following 40 epochs.

\section{Experiments}
\label{sec:experiments}

\begin{figure*}[t]
	\centering
	\includegraphics[width=\linewidth]{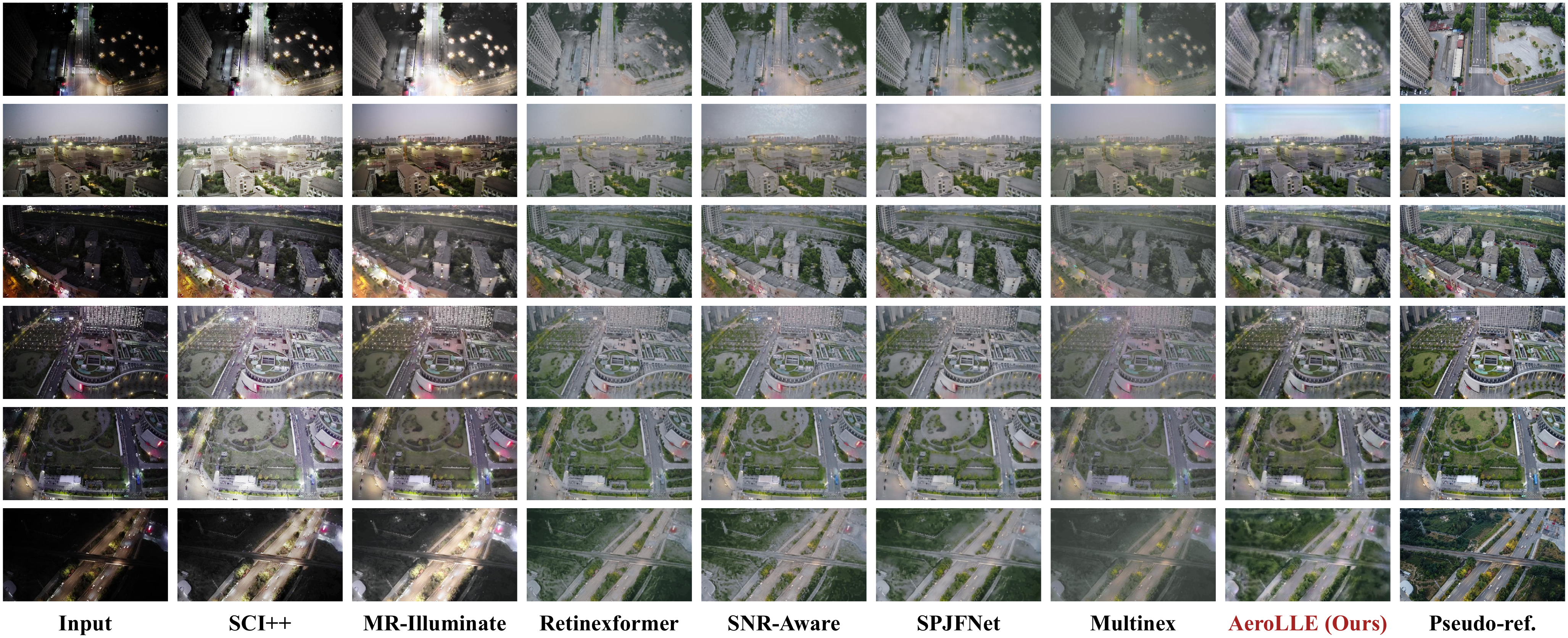}
	\caption{Qualitative comparison on six nighttime aerial scenes. Rows 1-3 are from Test 1 and rows 4-6 are from Test 2. The examples show the joint challenge of revealing dark surroundings and balancing nearby illuminated regions.}
	\label{fig:qualitative}
\end{figure*}

\subsection{Experimental Setup}

\textbf{Training and validation.}
AeroLLE uses 900 training pairs and 100 validation pairs
for checkpoint selection.
Both stages are trained from scratch.
The Base Enhancer is trained for 2,000 epochs using Adam
with batch size 8 and 384$\times$384 crops accompanied
by full-scene thumbnails.
Its learning rate follows a cosine schedule from
3$\times 10^{-4}$ to $10^{-4}$.
The checkpoint with the highest validation PSNR is selected
and frozen for SAECC training.
SAECC is trained on full-resolution images for 300 epochs
using AdamW with batch size 2.
Its learning rate ranges from 4$\times$$10^{-5}$
to 2$\times$$10^{-6}$, with a 10-epoch warm-up.
The epoch-300 checkpoint is used for evaluation.
AeroLLE and the comparison methods use the same
RTX 5090 GPU configuration.

\textbf{Comparison methods.}
We compare SCI++~\cite{SCIpp}, Zero-DCE~\cite{ZeroDCE},
SCI~\cite{SCI}, ZeroIG~\cite{ZeroIG},
MR.~Illuminate~\cite{MR}, ZeroIDIR~\cite{ZeroIDIR},
CIDNet~\cite{HVI}, Retinexformer~\cite{Retinexformer},
SNR-Aware~\cite{SNRAware}, SPJFNet~\cite{SPJFNet},
and Multinex~\cite{Multinex} using their official implementations.
Comparison methods trained on AeroNight use the 1,000-pair
development pool and follow their official configurations
for training and checkpoint selection.

\textbf{Metrics and complexity.}
PSNR and SSIM~\cite{SSIM} measure agreement with the screened
pseudo-references on Test~1 and Test~2.
SSIM uses a \(7\times7\) uniform window.
Parameter counts and FLOPs cover both inference stages,
with FLOPs measured at \(512\times512\).
FPS is measured on the RTX 5090 with full-resolution
\(2736\times1536\) inputs and batch size one.

\subsection{Quantitative Comparison}

\begin{table}[t]
	\centering
	\caption{Quantitative comparison on the two 100-pair AeroNight test sets. PSNR (dB) and SSIM are computed against screened pseudo-references. FLOPs are measured at \(512\times512\). FPS is measured at \(2736\times1536\) with batch size one on an RTX 5090.}
	\label{tab:comparison}
	\begingroup
	\setlength{\tabcolsep}{1.6pt}
	\renewcommand{\arraystretch}{1.2}
	\fontsize{6.5}{7.5}\selectfont
	\resizebox{\columnwidth}{!}{%
		\begin{tabular}{@{}clrrrrrrr@{}}
			\toprule
			\multirow{2}{*}{} & \multirow{2}{*}[-2.5pt]{Method} & \multirow{2}{*}[-2.5pt]{Params $\downarrow$} & \multirow{2}{*}[-2.5pt]{FLOPs $\downarrow$} & \multirow{2}{*}[-2.5pt]{\shortstack{FPS $\uparrow$\\(img/s)}} & \multicolumn{2}{c}{Test 1} & \multicolumn{2}{c}{Test 2} \\
			\cmidrule(lr){6-7}\cmidrule(l){8-9}
			& & & & & PSNR $\uparrow$ & SSIM $\uparrow$ & PSNR $\uparrow$ & SSIM $\uparrow$ \\
			\midrule
			\multirow{6}{*}{\rotatebox[origin=c]{90}{\scriptsize Unsupervised}} & SCI++~\cite{SCIpp} & \textbf{0.25K} & \textbf{0.13G} & \textbf{558.38} & 11.3936 & 0.2116 & 12.0601 & 0.2117 \\
			& Zero-DCE~\cite{ZeroDCE} & 79.42K & 41.52G & 25.37 & 10.3662 & 0.2518 & 11.0125 & 0.2583 \\
			& SCI~\cite{SCI} & \underline{0.26K} & \textbf{0.13G} & \underline{327.55} & 10.7230 & 0.1951 & 11.7967 & 0.2017 \\
			& ZeroIG~\cite{ZeroIG} & 62.85K & 71.42G & 12.95 & 11.5596 & 0.2021 & 12.2945 & 0.2040 \\
			& MR. Illuminate~\cite{MR} & 1.08B & 44.93T & 0.011 & 12.8490 & 0.3523 & 14.4819 & 0.3576 \\
			& ZeroIDIR~\cite{ZeroIDIR} & 141.42M & 72.50T & 0.042 & 11.9002 & 0.0896 & 12.7364 & 0.0961 \\
			\midrule
			\multirow{5}{*}{\rotatebox[origin=c]{90}{\scriptsize Supervised}} & CIDNet~\cite{HVI} & 1.98M & 65.43G & 3.11 & \underline{16.4094} & \underline{0.5154} & \underline{17.2551} & 0.4867 \\
			& Retinexformer~\cite{Retinexformer} & 1.61M & 133.25G & 2.69 & 16.0217 & 0.5040 & 16.9859 & 0.4856 \\
			& SNR-Aware~\cite{SNRAware} & 39.12M & 204.35G & 1.98 & 15.3914 & 0.4741 & 16.3511 & 0.4611 \\
			& SPJFNet~\cite{SPJFNet} & 917.37K & 74.82G & 4.68 & 16.0247 & 0.5153 & 17.0602 & \underline{0.4957} \\
			& Multinex~\cite{Multinex} & 44.62K & 16.88G & 1.45 & 15.3577 & 0.4934 & 16.3330 & 0.4696 \\
			\midrule
			& \textbf{AeroLLE (Ours)} & 255.24K & \underline{2.47G} & 62.12 & \textbf{17.6243} & \textbf{0.5190} & \textbf{17.8912} & \textbf{0.4964} \\
			\bottomrule
	\end{tabular}}
	\endgroup
\end{table}

\textbf{Enhancement quality.}
AeroLLE achieves the highest PSNR and SSIM on both test sets
in Table~\ref{tab:comparison}.
It reaches 17.62/17.89~dB PSNR and 0.5190/0.4964 SSIM
on Test~1/Test~2.
The PSNR gains over CIDNet, the strongest PSNR baseline,
are 1.21 and 0.64~dB.
AeroLLE also exceeds Retinexformer by 1.60 and 0.91~dB
and Multinex by 2.27 and 1.56~dB.

The supervised baselines outperform the reference-free and zero-shot methods on both metrics. Their results support the use of generated references as training targets across several enhancement architectures. AeroLLE improves on these trained baselines with a compact model, maintaining its advantage on both held-out test sets.

\textbf{Inference efficiency.}
AeroLLE uses 255.24K parameters and 2.47G FLOPs,
the lowest FLOP count among the compared supervised methods.
It reduces FLOPs by 85.4\% relative to Multinex
and processes full-resolution images at 62.12 FPS,
42.84 times the Multinex throughput.

The computational savings reflect the spatial organization of the model. The full-resolution path handles local image content while scene-context processing operates on a thumbnail and SAECC predicts calibration fields on a stride-32 grid. This allocation keeps image-wide conditioning and photometric refinement compact as input resolution grows. AeroLLE combines higher reference agreement with high throughput on native-resolution aerial images.

\subsection{Qualitative Comparison}

Figure \ref{fig:qualitative} illustrates the challenge of coordinating exposure across different regions of an aerial scene. In the first row SCI++ and MR. Illuminate brighten the road and existing lights while much of the surrounding vegetation and buildings remains dark. AeroLLE makes these surrounding regions easier to distinguish. The last row presents the same issue around a pedestrian bridge, where AeroLLE reveals the bridge and adjacent vegetation alongside the illuminated road to produce a more balanced view of the scene.

The supervised methods enhance broader areas while differing in regional contrast and color. In the last row Multinex produces pale vegetation and a low-contrast road surface. AeroLLE gives the vegetation stronger green tones and a clearer separation from the road. In the second row its sky, roofs, and vegetation also show more distinct color and brightness. These examples illustrate the value of jointly handling dark-region visibility and regional photometric balance in high-resolution aerial images.

\begin{figure*}[t]
	\centering
	\includegraphics[width=\linewidth]{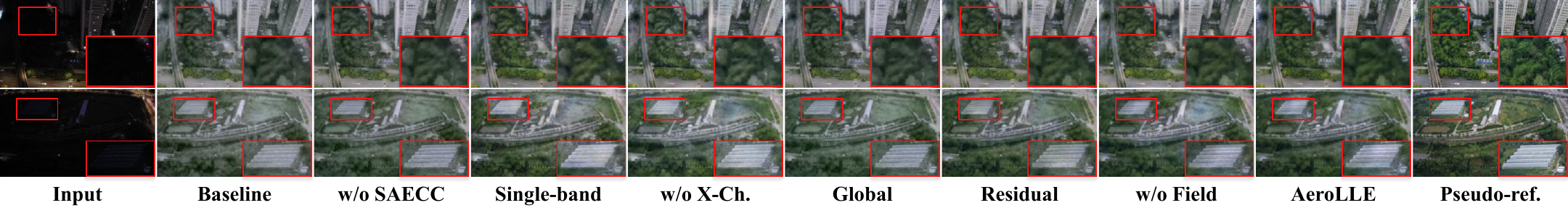}
	\caption{Qualitative ablation on two nighttime aerial scenes. Insets show vegetation (top) and a rooftop (bottom). Baseline is the Stage-1 variant without CSC; \mbox{w/o SAECC} denotes Stage-1 CSC. Red boxes mark insets; the final column is the screened pseudo-reference.}
	\label{fig:ablation_overall}
\end{figure*}

\begin{figure}[t]
	\centering
	\includegraphics[width=\linewidth]{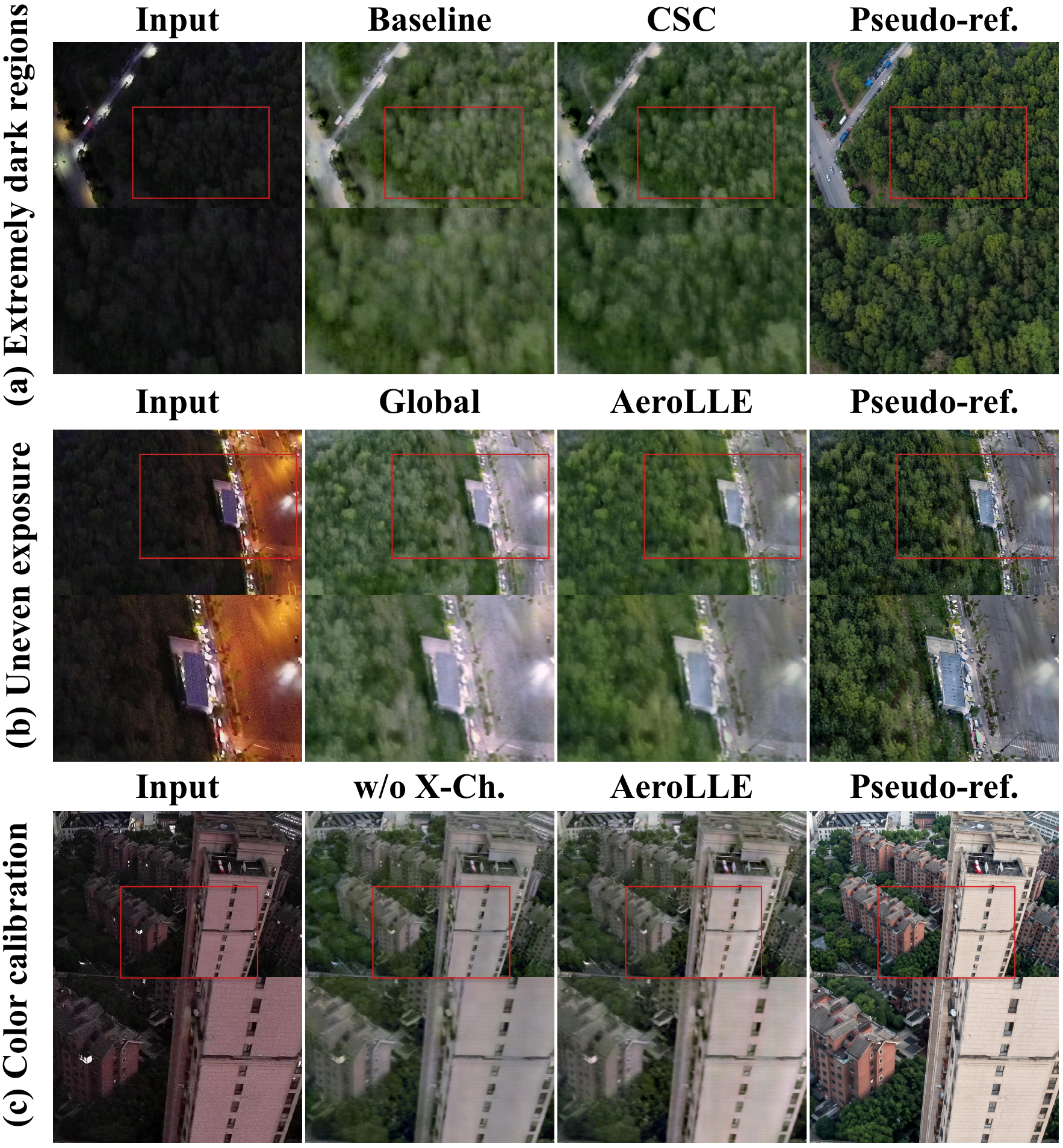}
	\caption{Component effects on regional exposure and color. (a) Baseline and CSC on dark vegetation. (b) Global affine calibration and AeroLLE on vegetation beside an illuminated road. (c) \mbox{w/o X-Ch.} and AeroLLE on building color. Lower rows in (b,c) enlarge the marked regions. Baseline and CSC are Stage-1 outputs. The last column shows the screened pseudo-reference.}
	\label{fig:ablation_challenge}
\end{figure}

\subsection{Ablation Studies}
\label{sec:ablation}

\begin{table}[t]
	\centering
	\caption{Ablation on two 100-pair AeroNight test sets; PSNR (dB) and SSIM use screened pseudo-references. The upper block compares Baseline without CSC with Stage-1 CSC; the lower block tests context and calibration variants. CSC and \mbox{w/o SAECC} are identical Stage-1 outputs.}
	\label{tab:ablation}
	\includegraphics[width=\columnwidth]{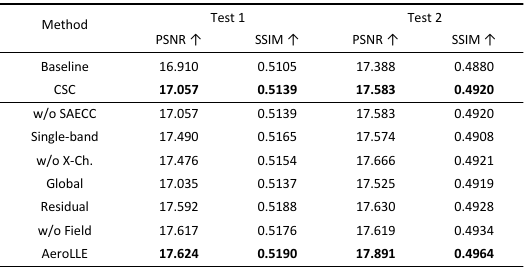}
\end{table}

Table~\ref{tab:ablation} compares Baseline, the Stage-1 model
without CSC, with CSC and full-framework variants.
Single-band evaluates single-band context processing,
and \mbox{w/o X-Ch.} removes latent cross-channel exchange.
These two variants use independently trained first and
second stages.
Global replaces spatial calibration with a global affine
transform, and Residual uses RGB residual prediction.
The \mbox{w/o Field} variant removes teacher-field supervision
and retains teacher-image supervision.
Figure~\ref{fig:ablation_overall} compares the variants
on vegetation and rooftop examples.

\textbf{Frequency-dependent context.}
CSC exceeds Baseline by 0.15~dB PSNR on Test~1
and 0.20~dB on Test~2, with SSIM increasing on both sets.
The complete model improves over Single-band by
0.13 and 0.32~dB.
Removing cross-channel exchange reduces PSNR by
0.15 and 0.23~dB.
Frequency partitioning and channel interaction both
contribute to the final enhancement quality.

These comparisons support frequency-dependent channel interactions, which allow the thumbnail representation to use different transforms for illumination variations at different scales. The effect is visible in Fig. \ref{fig:ablation_challenge}(a), where CSC produces deeper green tones and more distinct variation across the vegetation canopy. Panel (c) shows the associated color change in the complete model. AeroLLE produces a warmer building facade than \mbox{w/o X-Ch.}, closer to the pseudo-reference appearance.

\textbf{Spatial calibration.}
SAECC adds 0.57~dB on Test~1 and 0.31~dB on Test~2
over the Stage-1 output.
SSIM increases by 0.0051 and 0.0044.
The global affine variant scores below the Stage-1 output
on both sets, and spatial calibration exceeds Global
by 0.59 and 0.37~dB.

The comparison highlights the role of spatial variation in the remaining correction. The global affine variant uses a single mapping for the whole image and the spatial gain and bias fields let SAECC adjust each region. In Fig. \ref{fig:ablation_challenge}(b) Global produces a pale appearance across both the vegetation and road. AeroLLE lowers the road brightness and gives the vegetation a darker, more distinct green appearance. These observations connect the quantitative gain with differences in regional exposure and color balance within the same scene.

\textbf{Calibration representation.}
AeroLLE exceeds the RGB-residual variant by 0.03 dB on Test 1 and 0.26 dB on Test 2. The larger gap on the unseen-location set supports the use of explicit RGB gain and bias for spatial refinement in this setting. The fields express the remaining correction through regional exposure and channel-balance adjustments around the base estimate.

\textbf{Teacher-field supervision.}
Teacher-field supervision improves PSNR by 0.01 dB on Test 1 and 0.27 dB on Test 2. The corresponding SSIM gains are 0.0014 and 0.0030. Its larger benefit on Test 2 supports the value of explicit calibration targets at previously unseen locations.

The field losses supervise the gain and bias predicted by SAECC on its coarse grid. They complement teacher-image supervision with targets for the calibration parameters themselves. After bilinear upsampling the predicted fields determine regional exposure and color adjustments on the full-resolution base estimate.

\subsection{Discussion}

Generated-reference supervision offers a path toward expanding nighttime aerial datasets across locations and illumination conditions. Future research could couple reference generation with enhancement learning, using content consistency and photometric quality to guide automatic reference screening.
AeroNight provides a starting point for studying how reference quality and dataset diversity affect enhancement at unseen locations, including how supervision can scale as new flight data become available.

\section{Conclusion}
\label{sec:conclusion}
AeroNight addresses the supervision bottleneck in nighttime aerial
enhancement by pairing 1,200 real nighttime UAV images with screened,
generated normal-light references.
This approach retains real nighttime degradation in the inputs
and makes captured observations available for supervised learning.
AeroLLE combines scene-level frequency context with bounded spatial
exposure--color calibration, achieving the highest PSNR and SSIM
among the evaluated methods on both test sets with 255.24K parameters
and 2.47G FLOPs.
The gains from spatial calibration and teacher-field supervision
support regional photometric correction and field-level targets for enhancement at unseen locations.
AeroNight provides the community with a shared benchmark for
studying reference generation and enhancement together under
real aerial capture conditions.

\appendix
\section*{Supplementary Overview}

This supplement provides the evidence and implementation details that complement the main paper. It first documents how AeroNight was constructed and why its pseudo-reference source was selected, then specifies the two-stage AeroLLE pipeline and its training and evaluation settings. Additional experiments examine transfer to other aerial datasets, dataset-specific training on three LOL benchmarks, random-seed variation, and reference-free image statistics. The supplement also provides the core training and inference procedure and further AeroNight test-set comparisons.

\hyperref[sec:dataset]{Dataset Preparation} describes field collection, candidate generation, screening, data splits, and photometric descriptors. \hyperref[sec:reference]{Reference-Source Examples} compares generated candidates and explains the choice of Gemini 3.1 Pro. \hyperref[sec:formulation]{Detailed Formulation} and \hyperref[sec:procedure]{Core Training and Inference Procedure} specify the model operations and training path. \hyperref[sec:training]{Training and Evaluation Details} gives the optimization, image interface, and metric conventions. \hyperref[sec:datasets]{Additional Dataset Comparisons} presents public-dataset context and aerial and non-aerial results. \hyperref[sec:quantitative]{Additional Quantitative Analyses} reports the second random-seed run and reference-free measures. \hyperref[sec:qualitative]{Additional Qualitative Comparisons} extends the visual results on both AeroNight test sets.

Throughout the supplement, a pseudo-reference denotes a generated normal-light image retained after manual content and alignment screening. It supplies appearance guidance for exposure and color correction, while the captured nighttime input supplies scene structure and real nighttime degradation. We use \(x\) for the captured nighttime image, \(y_p\) for its screened appearance reference, \(B_0\) for the quantized base estimate, and \(B_1\) for the calibrated output.

\section{Dataset Preparation}\label{sec:dataset}

\subsection{Field collection and platform}

Nighttime aerial videos were captured at multiple sites and aerial viewpoints in the urban areas of Hefei and Lu'an, Anhui, China, with UAV flight altitudes ranging from 30 to 120 m. The flights used a UAV equipped with a Sony camera with a resolution of \(7680\times4320\) pixels. All sites were outside areas where filming is prohibited. Images containing faces, license plates, or other privacy-sensitive content were excluded during screening. The recorded videos were sampled at five-second intervals, producing more than 1,600 candidate frames. Frames with high visual similarity, severe blur, or unrecognizable content were removed before pseudo-reference generation.

The retained inputs contain the conditions targeted by AeroLLE, including spatially nonuniform exposure, mixed color temperatures, localized saturated lights, sensor noise, and compression artifacts. The image pairs used by the benchmark are represented at \(2736\times1536\) resolution after the source-frame preparation step.

\begin{figure*}[t]
\centering
\includegraphics[width=0.77\textwidth]{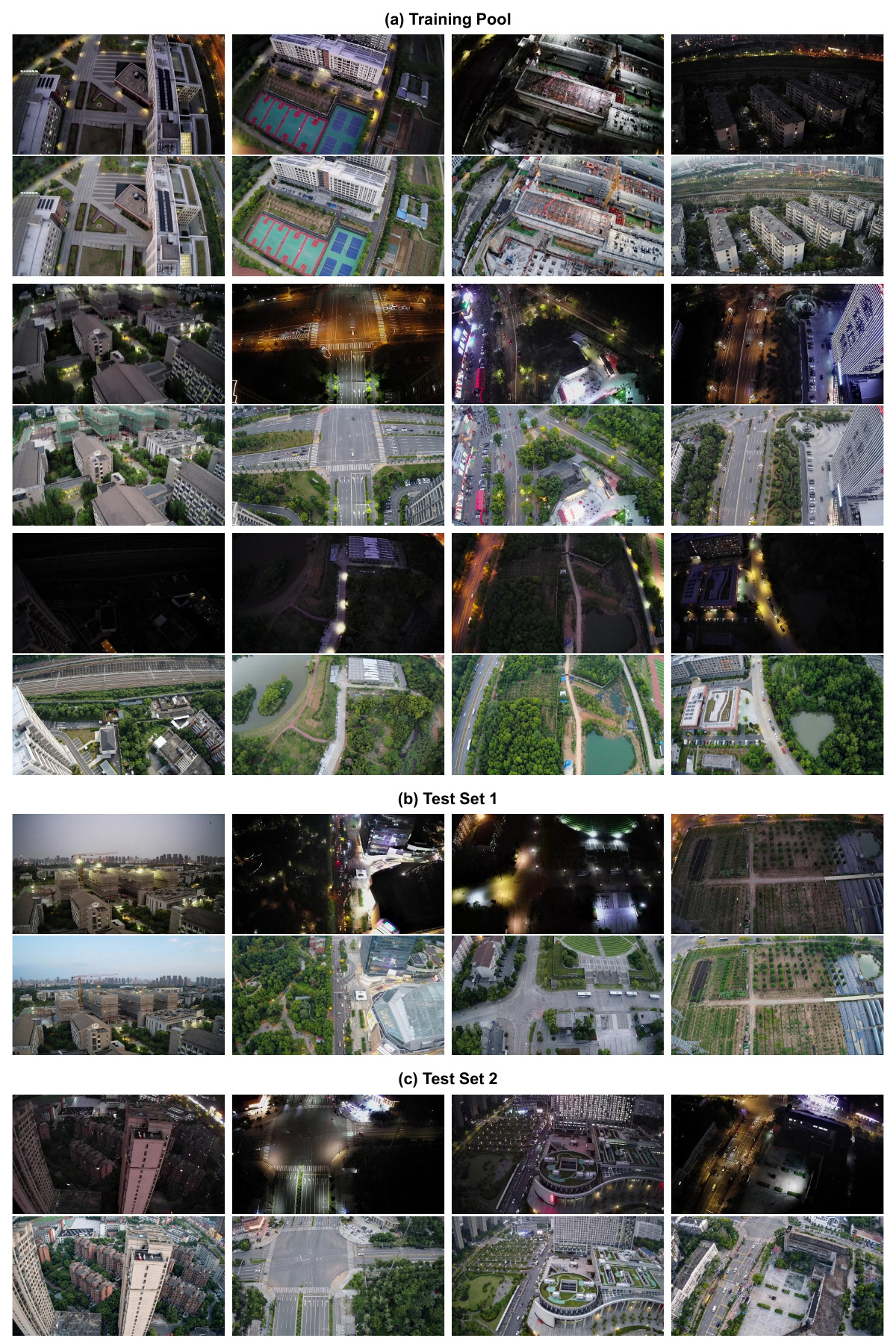}
\caption{Representative AeroNight pairs. The upper and lower images in each pair show the captured nighttime input and its screened generated normal-light pseudo-reference, respectively. The display contains 12 pairs from the training pool, four pairs from Test 1, and four pairs from Test 2.}
\label{fig:s2}
\end{figure*}

\subsection{Pseudo-reference generation}

Capturing a registered normal-light target from a moving UAV is difficult because a later flight can differ in viewpoint, traffic, shadows, local illumination, and atmospheric conditions. We therefore generate a normal-light candidate from each retained nighttime observation. Synthesizing a dark input from a normal-light photograph can weaken or remove sensor noise and other camera-processing traces already altered by denoising, tone mapping, or compression. Starting from the captured nighttime frame preserves the degradation encountered by the enhancement model in practice.

We use Gemini 3.1 Pro to generate the normal-light candidates. The generation instruction is: ``Generate a normal-light daytime image from this nighttime image. Preserve pixel-level alignment with the input. Do not change any visible structures, buildings, vehicles, roads, vegetation, or their spatial relationships. Do not add shadows or alter scene content. Change only the illumination and color conditions. If text is unclear, remove it rather than inventing characters. The output should support pixel-level comparison with the input.''

A compact enhancement model is first applied to the original nighttime image to provide a clearer view during screening. The collectors know the captured sites and use this view together with the original frame to inspect candidate content. Two to four generations are compared for each input and the most suitable candidate is retained. If obvious hallucinations remain after five to eight generations, the input is removed from the benchmark.

Image generation uses a closed API, so the same prompt may not reproduce identical outputs. We will release the finalized AeroNight dataset as a fixed collection of nighttime inputs and screened pseudo-references for future research.

\subsection{Screening, reference role, and split design}\label{sec:screening}

Candidates are manually compared with their nighttime inputs. Outputs that add or remove visible objects, rewrite text or road markings, change road or roof boundaries, introduce geometric misalignment, or produce implausible illumination or color are removed. Generated images can contain hallucinated content. Manual screening reduces these artifacts as much as possible, although it cannot guarantee their complete removal. We therefore refer to the retained generated images as pseudo-references. The screened reference is used for appearance guidance in color correction, channel balance, and exposure adjustment. It is not used as structural ground truth. The captured nighttime image supplies the observed scene structure, while the generated reference supplies the normal-light appearance direction.

AeroNight contains 1,200 pseudo-paired samples. We use 900 pairs for training and 100 pairs for internal model selection. Test 1 contains 100 pairs of different images captured at locations represented in the development data, while Test 2 contains 100 pairs captured at locations not represented in development. Both test sets are evaluated after training and internal checkpoint selection.

\subsection{Dataset photometric descriptors}

Each image is area-downsampled with preserved aspect ratio to a maximum side length of 512 pixels, and RGB values are normalized to \([0,1]\). Luminance is \(Y=0.299R+0.587G+0.114B\). Mean luminance and RMS contrast are its pixel mean and population standard deviation. Luminance range is \(P_{95}(Y)-P_{05}(Y)\); spatial luminance range uses the same percentile difference on an \(8\times8\) area-averaged luminance grid. Shadow and highlight occupancy are the pixel fractions with \(Y<0.10\) and \(Y>0.90\), respectively.

Mean saturation averages \((\max\mathrm{RGB}-\min\mathrm{RGB})/\max\mathrm{RGB}\) over pixels, with zero assigned to black pixels. For \(r=R-G\) and \(b=(R+G)/2-B\), colorfulness is \(\sqrt{\operatorname{Var}(r)+\operatorname{Var}(b)}+0.3\sqrt{\operatorname{mean}(r)^2+\operatorname{mean}(b)^2}\). Each luminance KDE is normalized to its own peak, and the bars in the contrast and occupancy panels span the first to third quartiles. Paired changes are computed as reference minus input for each descriptor. The confidence intervals in the main-paper analysis use 10,000 image-pair bootstrap resamples with seed 3415.

\section{Reference-Source Examples}\label{sec:reference}

\begin{figure*}[!t]
\centering
\includegraphics[width=\textwidth]{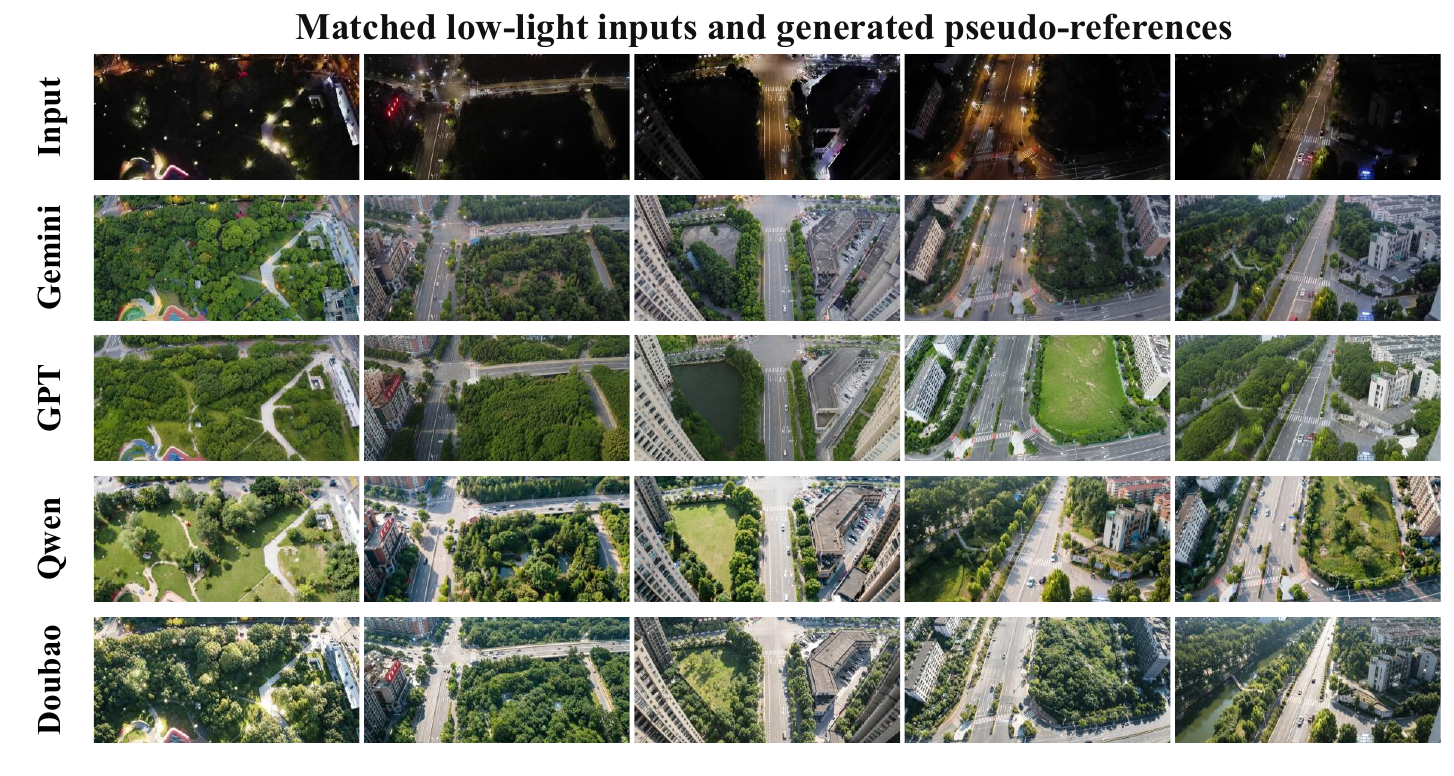}
\caption{Pseudo-reference source examples, part I. Each column corresponds to one nighttime aerial input. Rows show Input, Gemini, GPT, Qwen, and Doubao in that order.}
\label{fig:s3}

\vspace{6pt}

\centering
\includegraphics[width=\textwidth]{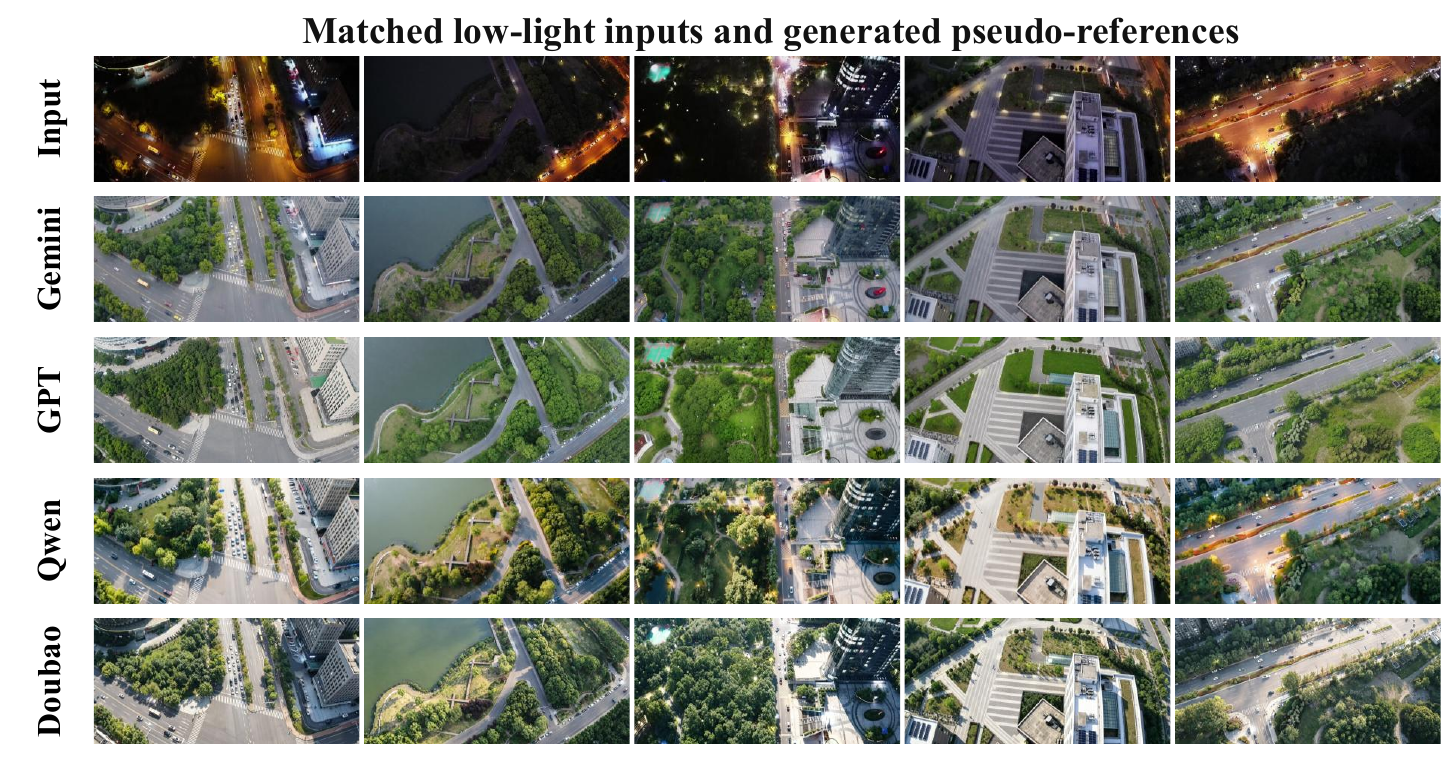}
\caption{Pseudo-reference source examples, part II. The row order follows Figure~\ref{fig:s3}. Additional scenes show exposure and color changes across the candidate sources.}
\label{fig:s4}
\vspace{32pt}
\end{figure*}

We compared four image-generation sources on the same 20 nighttime aerial inputs. The comparison includes Gemini 3.1 Pro (Gemini), OpenAI GPT Image 2 (GPT), Qwen-Image-2.0 (Qwen), and Doubao Seedream 4.5 (Doubao). Figures~\ref{fig:s3} and~\ref{fig:s4} show ten representative inputs and their generated images. All four sources increase daylight-like visibility, but their scene-dependent changes differ around intersections, parked vehicles, building edges, vegetation, and nearly black regions. Gemini produces a consistent global exposure and color response across the displayed cases. Several outputs from the other sources show larger changes in road boundaries, object counts, vegetation texture, or shadows. In particular, Gemini keeps roads and buildings recognizable while producing a more even daylight-like appearance around the illuminated regions. These observations motivate a quantitative comparison of appearance and consistency with the input.

Table~\ref{tab:s1} averages image-level descriptors over the 20 inputs. NIQE, BRISQUE, and IL-NIQE summarize natural-image statistics. Lightness order error (LOE), highlight clipping (HC), and gray-world deviation (GWD) describe exposure and color. Edge F1, added-edge (AE) and missing-edge (ME) fractions, input SSIM (SSIM-L), and SIFT coverage (SIFT-C) describe complementary aspects of consistency with the input. Gemini obtains NIQE of 2.345, LOE of 0.261, HC of 0.049\%, and GWD of 0.045, the lowest values in their respective columns. It also achieves the highest Edge F1 (0.539), SSIM-L (0.358), and SIFT-C (0.613), together with the lowest added-edge fraction (0.510) and missing-edge fraction (0.315). These results indicate a favorable balance between natural appearance, illumination correction, and agreement with structures visible in the low-light inputs. Doubao obtains the lowest BRISQUE and Qwen the lowest IL-NIQE. Taken together with the visual comparison, these results led us to select Gemini 3.1 Pro as the source of AeroNight pseudo-references. The generated candidates were subsequently screened using the \hyperref[sec:screening]{screening protocol}.

\begin{table*}[t]
\centering
\includegraphics[width=\textwidth]{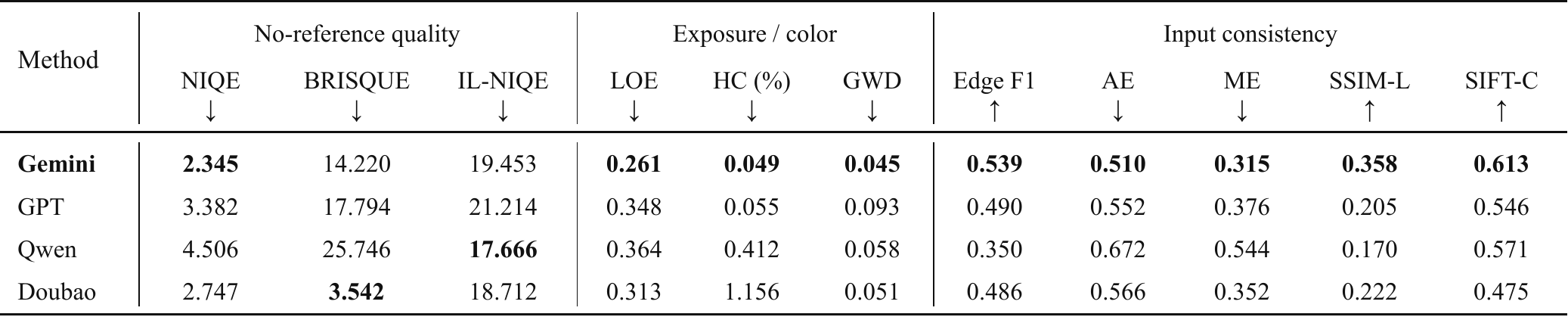}
\caption{Comparison of visual image-generation models used as candidate pseudo-reference sources. The table reports no-reference quality, exposure and color descriptors, and input-consistency measures on the same nighttime aerial inputs. HC denotes highlight clipping, GWD gray-world deviation, AE and ME added- and missing-edge fractions, SSIM-L SSIM to the low-light input, and SIFT-C the spatial coverage of geometrically consistent matches. Gemini 3.1 Pro is selected for AeroNight based on its appearance and consistency results.}
\label{tab:s1}
\end{table*}

\section{Detailed Formulation}\label{sec:formulation}

\subsection{Thumbnail encoder and residual latent \mbox{filtering}}

The thumbnail encoder supplies scene-wide illumination information at low spatial cost. The input is bilinearly resized with preserved aspect ratio to a maximum side length of 384 pixels. Four \(3\times3\) convolutions use channel widths \(3\!\to\!16\!\to\!24\!\to\!24\!\to\!24\). The first two have stride 2, and the last two use stride 1 with dilation 2 and 4. Each convolution is followed by channel-wise PReLU. The resulting feature grid is approximately one-quarter of the thumbnail size along each dimension.

Before CSC, a \(1\times1\) projection reduces the 24-channel features to eight latent channels. Each channel is filtered independently with a learned \(5\times5\) kernel normalized by softmax over its 25 entries. Channel mixing is provided by the input and output projections, and by the subsequent CSC operator. For the Fourier response \(\widehat K_c\) of a normalized kernel and \(\lambda_c=0.25+0.75\,\operatorname{sigmoid}(\ell_c)\), the forward-filter response is

\begin{equation}
\begin{aligned}
H_c(\omega)&=\frac{\widehat K_c(\omega)+\lambda_c}{1+\lambda_c},\\
V'_c&=\operatorname{IRFFT2}\!\left(H_c\odot\operatorname{RFFT2}(V_c)\right).
\end{aligned}
\label{eq:s1}
\end{equation}

The filtered channels pass through GELU and a \(1\times1\) output projection to 24 channels, followed by residual addition. The output projection is initialized to zero. Filtering uses periodic boundaries, with centered kernel offsets folded onto the feature grid before the Fourier transform. This operator filters latent scene features before frequency-dependent channel exchange. The softmax normalization makes each spatial kernel nonnegative with unit sum. With \(\lambda_c\in[0.25,1]\), Eq. \eqref{eq:s1} is a convex mixture of that filter and the identity, giving unit response at zero frequency and magnitude at most one for the filter response. The identity weight \(\lambda_c/(1+\lambda_c)\) lies in \([0.2,0.5]\), retaining an unfiltered component within the latent filtering operation before the output projection.

\subsection{CSC masks, transfers, and context readout}

Let \(f_y\) and \(f_x\) be normalized frequencies from FFTFreq and RFFTFreq, and let \(r^2=f_y^2+f_x^2\). Three radial responses are defined as \(q_\sigma=\exp(-r^2/\sigma^2)\), with \(\sigma=0.025\), \(0.075\), and \(0.20\). The four fixed masks are

\begin{equation}
\begin{aligned}
m_1&=q_{0.025},&m_2&=q_{0.075}-q_{0.025},\\
m_3&=q_{0.20}-q_{0.075},&m_4&=1-q_{0.20}.
\end{aligned}
\label{eq:s2}
\end{equation}

The masks are nonnegative and sum to one at each frequency. The first emphasizes low-frequency context, the middle two cover progressively higher frequencies, and the last retains the remaining high-frequency response. Each band applies its own bounded, off-diagonal \(24\times24\) real transfer matrix to the complex spectrum. The four masked responses are summed, transformed back with the original spatial size, and added to the input features. RFFT2 and IRFFT2 use orthonormal normalization. The diagonal coefficients are masked out; the registered matrices contain 2,304 coefficients, of which 2,208 are off-diagonal entries. The three scales give smooth transitions from broad spatial variation to finer context responses, and the fourth mask completes the spectral coverage. For \(C=24\), the factor \(0.12/\sqrt C\) limits each off-diagonal transfer coefficient to approximately 0.0245 in magnitude; zero initialization starts the exchange at the unchanged context.

The shared context features drive both local feature modulation and spatial input correction. One \(1\times1\) head predicts six maps, split into RGB gain-related and bias-related maps. A second \(1\times1\) head follows global average pooling and sigmoid to produce a 12-dimensional scene-channel attention vector. Both heads have zero-initialized weights and biases, giving zero spatial maps and an initial attention value of 0.5. At inference, the maps are bilinearly resized to the input resolution; during crop training, they are sampled at the corresponding full-scene crop coordinates.

\subsection{Local feature modulation and first-stage \mbox{objective}}

The local feature modulation extends the compact MobileIE backbone \cite{MobileIE}. Let \(V\) denote the 12-channel output of the local body and \(a_c\) the scene-channel attention. The spatial attention is \(a_s=\operatorname{sigmoid}(\operatorname{Conv}_{1\times1}(\max_{\mathrm{channel}}(a_c\odot V)))\). The modulated feature is \(a_c\odot a_s\odot V\). The model then multiplies this feature by a luminance and local-boundary gate. With \(Y=0.299R+0.587G+0.114B\) and \(P_3\) a zero-padded \(3\times3\) average pool, the gate is

\begin{equation}
\begin{aligned}
\eta=\operatorname{clip}\big(&1+u_s(1-Y)\\
&+v|Y-P_3(Y)|,\,0.5,\,1.5\big).
\end{aligned}
\label{eq:s3}
\end{equation}

\begin{equation}
\begin{aligned}
u_s=u\big[1+\tanh(\gamma)\operatorname{clip}\big(&0.18-\operatorname{mean}(Y),\\
&-0.18,0.18\big)\big].
\end{aligned}
\label{eq:s4}
\end{equation}

The learned scalars \(u\) and \(v\) start at 0.1, and \(\gamma\) starts at zero. The RGB tail reads \(\eta\)-modulated features and predicts the correction used in main-text Eqs. (4)--(5). The scene statistic in this modulation is computed on the currently processed local input. The context attention and maps, in contrast, originate from the full-scene thumbnail. The gate interval \([0.5,1.5]\) permits attenuation and amplification of local features around unit scaling while bounding this multiplicative modulation.

\begin{figure*}[!t]
\centering
\includegraphics[width=0.94\textwidth]{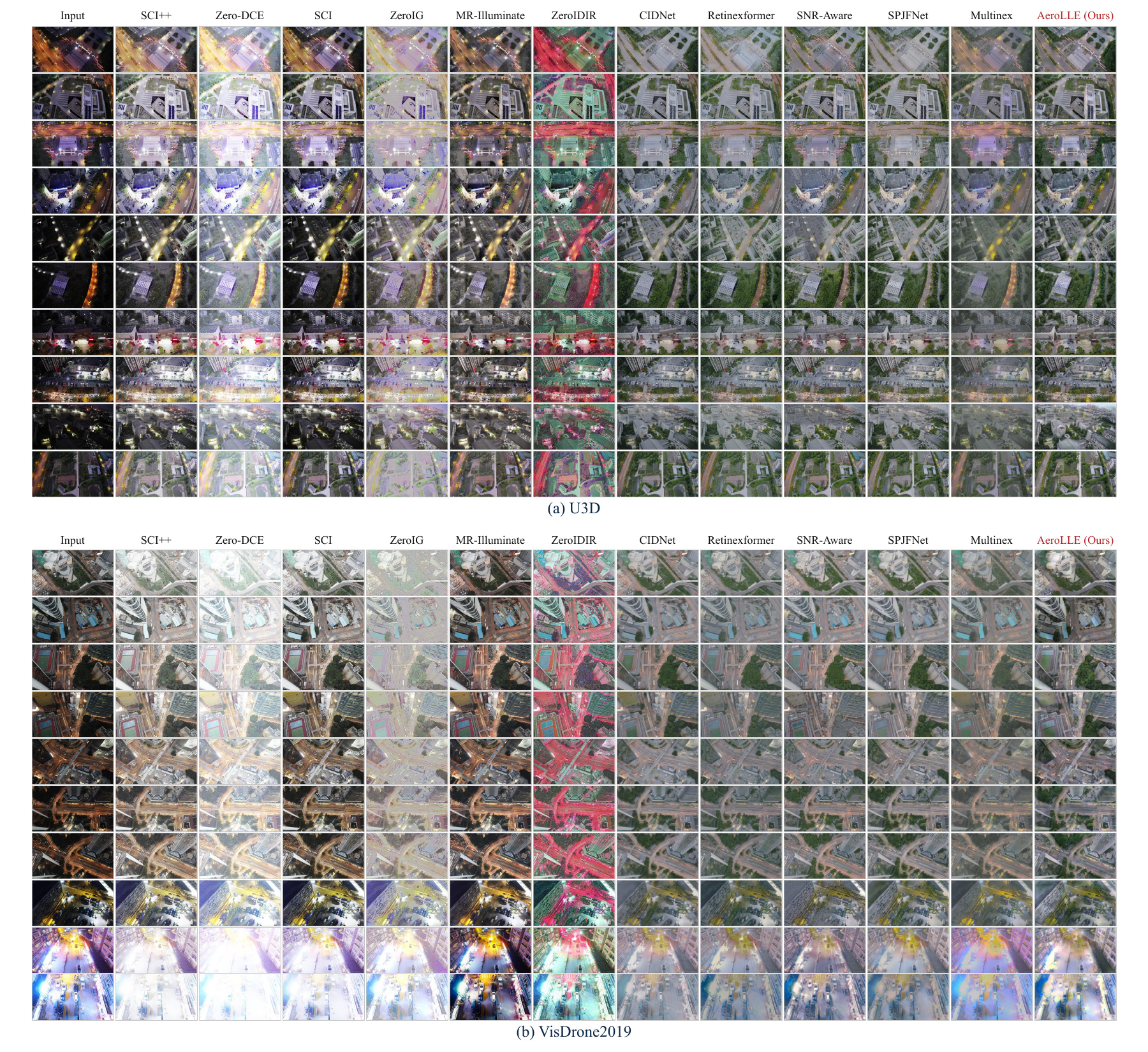}
\caption{Qualitative transfer to additional aerial datasets. The model is trained on AeroNight and directly applied to U3D and VisDrone2019. Each row lists the input, comparison methods, and AeroLLE output.}
\label{fig:s9}
\end{figure*}

The first-stage appearance objective uses the unclipped network prediction \(p=F_\theta(x)\) and target \(t=y_p\). Define \(d=p-t\), \(\Delta_x\) and \(\Delta_y\) as first-order finite differences along the two spatial axes, and mean as averaging over the indicated tensor entries. The objective is

\begin{equation}
\begin{aligned}
L_1={}&\operatorname{mean}(d^2)+0.10\operatorname{mean}(|d|)\\
&+0.05L_{\mathrm{grad}}+0.01L_{\mathrm{mean}}\\
&+0.01L_{\mathrm{range}}+0.10L_{\mathrm{under}}.
\end{aligned}
\label{eq:s5}
\end{equation}

The gradient term is \(L_{\mathrm{grad}}=\operatorname{mean}(|\Delta_xY(d)|)+\operatorname{mean}(|\Delta_yY(d)|)\). The channel-mean term is \(L_{\mathrm{mean}}=\operatorname{mean}(|\operatorname{mean}_{HW}(d)|)\), averaging the absolute spatial mean residual across RGB channels and the batch. The range term is \(L_{\mathrm{range}}=\operatorname{mean}(\operatorname{ReLU}(-p)^2)+\operatorname{mean}(\operatorname{ReLU}(p-1)^2)\). These terms respectively encourage appearance-gradient agreement, channel-balance agreement, and valid output range.

\begin{equation}
\begin{aligned}
L_{\mathrm{under}}=\operatorname{mean}\big(&m\odot\operatorname{ReLU}\big(Y(t)\\
&-Y(p)-0.02\big)^2\big).
\end{aligned}
\label{eq:s6}
\end{equation}

\begin{equation}
m=\operatorname{clip}\big(4Y(t)[1-Y(t)],\,0,\,1\big).
\label{eq:s7}
\end{equation}

The midtone weight \(m\) peaks at target luminance 0.5 and vanishes at 0 and 1, concentrating the underexposure term away from saturated highlights and near-black targets. The 0.02 margin is approximately five levels on an 8-bit luminance scale. The range loss penalizes predictions outside \([0,1]\), while the RGB and gradient terms fit reference appearance. The first-stage network is optimized with all terms in Eq. \eqref{eq:s5}.

\subsection{Thirteen-channel calibration conditions}

Let \(\bar x\) and \(\bar B_0\) be area-resized RGB images on the \(h\times w\) calibration grid, where \(h=\lceil H/32\rceil\) and \(w=\lceil W/32\rceil\). Their intensities are \(I_x=\max\mathrm{RGB}(\bar x)\) and \(I_0=\max\mathrm{RGB}(\bar B_0)\). Input darkness is \(D_x=1-I_x\). A normalized Sobel magnitude of \(Y(\bar x)\) gives the edge cue \(E_x\). The detail cues \(Q_x\) and \(Q_0\) are the mean absolute channel responses to a fixed Laplacian applied to \(\bar x\) and \(\bar B_0\), respectively. At \(2736\times1536\) input resolution, the calibration head operates on \(48\times86\) locations while the final affine transform remains at full resolution. The complete condition is

\begin{equation}
\begin{aligned}
C_{13}=\operatorname{Concat}\big[&\bar x(3),\bar B_0(3),I_x,I_0,D_x,\\
&E_x,Q_x,Q_0,r\big].
\end{aligned}
\label{eq:s8}
\end{equation}

\begin{equation}
r=\frac{\log\!\left(\operatorname{clip}\!\left[\frac{I_0+10^{-3}}{I_x+10^{-3}},0.25,8\right]\right)}{\log(8)}.
\label{eq:s9}
\end{equation}

The Sobel horizontal kernel is \(\left[\begin{smallmatrix}-1&0&1\\-2&0&2\\-1&0&1\end{smallmatrix}\right]/8\), and the vertical kernel is its transpose. The magnitude is \(\sqrt{g_x^2+g_y^2+10^{-8}}\), clipped to \([0,1]\). The Laplacian kernel is \(\left[\begin{smallmatrix}0&-1&0\\-1&4&-1\\0&-1&0\end{smallmatrix}\right]/4\). These filters use reflection padding. Intensity, gradient, detail, and ratio cues are all constructed after area resizing. The RGB channels supply coarse appearance information, while the scalar cues expose intensity relationships and local variation to the calibration head. The \(10^{-3}\) offset stabilizes the exposure ratio at near-black input intensities, and clipping the ratio to \([0.25,8]\) bounds its logarithmic cue to \([-2/3,1]\). The 13-channel condition therefore combines six RGB channels with seven scalar cues.

\subsection{Compact head and bounded field application}

The calibration head begins with a \(3\times3\) convolution from 13 to 64 channels and SiLU. Three residual blocks follow, each containing two \(3\times3\) convolutions, each followed by GroupNorm with eight groups and SiLU. The skip connection is the identity. A final \(3\times3\) convolution predicts six channels and is initialized to zero. The first three channels parameterize gain, and the remaining three parameterize bias according to main-text Eq. (6). Both fields are bilinearly interpolated with \texttt{align\_corners=False}.

In the final configuration, gain and bias are applied at full strength during both training and evaluation, corresponding to \texttt{field\_strength = 1.0}. Independent RGB fields allow exposure and channel balance to be corrected jointly: gain changes the local multiplicative response, while bias provides an additive offset. The fields are predicted on the stride-32 grid, bilinearly upsampled, and applied to the fixed base estimate before output clipping.

\section{Core Training and Inference Procedure}\label{sec:procedure}

The following pseudocode summarizes the two-stage training and inference path implemented by AeroLLE. Stage 1 learns the base enhancer and produces the quantized intermediate estimate. Stage 2 uses the nighttime input and this estimate to predict spatial gain and bias fields. The generated pseudo-reference is used during training and screening, but is not required at inference time.

The training path adds the screened pseudo-reference $y_p$ to the loss computation and forms an analytic teacher from $B_0$ and $y_p$. At inference, the network executes the same two stages using only $x$ and the quantized intermediate estimate.

\begin{figure*}[t]
\centering
\begin{minipage}{0.92\textwidth}
\small
\hrule\vspace{2pt}
\textbf{Algorithm 1. Core AeroLLE training and inference}\\[-1pt]
\hrule\vspace{2pt}
1. \textbf{Input:} Stage-1 training pairs $(x,y_p)$, Stage-1 parameters $\theta_1$, Stage-2 pairs, and field stride $s=32$.\\
2. \textbf{Output:} calibrated image $B_1$ and learned parameters $\theta_1,\theta_2$.\\
3. \textbf{for} each Stage-1 pair $(x,y_p)$ \textbf{do}\\
4. \quad Compute $p=F_{\theta_1}(x)$ and $m=\operatorname{clip}(4Y(y_p)[1-Y(y_p)],0,1)$.\\
5. \quad Evaluate $L_1$ in Eq.~\eqref{eq:s5} using the appearance, gradient, channel-mean, range, and\\
\hspace*{1.4em} midtone-weighted underexposure terms; update only $\theta_1$ with Adam.\\
6. \textbf{end for}; select $\theta_1$ on the internal split, freeze it, and cache $B_0=Q(\Pi(F_{\theta_1}(x)))$ for Stage 2.\\
7. \textbf{for} each Stage-2 pair $(x,y_p)$ at epoch $k$ \textbf{do}\\
8. \quad Set $(h,w)=(\lceil H/s\rceil,\lceil W/s\rceil)$ and area-resize $x$ and $B_0$; build\\
\hspace*{1.4em} $C_{13}=[\bar x,\bar B_0,I_x,I_0,D_x,E_x,Q_x,Q_0,r]$ using Eqs.~\eqref{eq:s8}--\eqref{eq:s9}.\\
9. \quad Split the six-channel head output into $(z_g,z_a)$, map $G_\ell=\exp(0.28\tanh(z_g))$ and\\
\hspace*{1.4em} $A_\ell=0.08\tanh(z_a)$, bilinearly upsample them to $G,A$, and compute $B_1=\Pi(B_0\odot G+A)$.\\
10. \quad Construct $g^*=\operatorname{clip}(\bar y_p/\max(\bar B_0,10^{-3}),0.7,1.6)$ and\\
\hspace*{1.4em} $a^*=\operatorname{clip}(\bar y_p-g^*\odot\bar B_0,-0.08,0.08)$, then form $B^*=\Pi(B_0\odot U(g^*)+U(a^*))$.\\
11. \quad Set $\rho(k)=0$ for $k\leq24$ and $\rho(k)=\min(1,(k-24)/40)$ otherwise; minimize the\\
\hspace*{1.4em} teacher-image $\ell_1$, teacher-field Smooth-$\ell_1$, total-variation, and $\rho(k)$-weighted reference and dark-region terms.\\
12. \quad Update only $\theta_2$ with AdamW; \textbf{end for}.\\
13. \textbf{Inference:} given only $x$, compute $B_0=Q(\Pi(F_{\theta_1}(x)))$ and execute Lines 8--9; $y_p$ and the\\
\hspace*{1.4em} analytic teacher are unused.\\
14. \textbf{return} $B_1$.
\vspace{2pt}\hrule
\end{minipage}
\end{figure*}

\subsection{Calibration magnitude}

For one channel and pixel, write \(B_0=b\) with \(b\in[0,1]\). The field parameterization and bilinear interpolation imply \(\exp(-0.28)\leq G\leq\exp(0.28)\) and \(|A|\leq0.08\). Hence,

\begin{equation}
\begin{aligned}
|bG+A-b|&\leq|G-1|+|A|\\
&\leq\exp(0.28)-1+0.08.
\end{aligned}
\label{eq:s10}
\end{equation}

Projection onto \([0,1]\) is non-expansive and leaves \(b\) unchanged, so the same bound holds for \(|\Pi(bG+A)-b|\). This bounds the second-stage photometric increment. The affine operator scales and offsets \(B_0\) at each fixed pixel location, while bilinear interpolation restricts field variation to the coarse-grid representation.

\subsection{Closed-form teacher and scheduled \mbox{supervision}}

The affine teacher expresses reference exposure and color on the same coarse grid used by SAECC. Let \(\bar t\) denote the area-resized reference and \(\bar B_0\) the area-resized, clipped base image. Their ratio and remaining residual define

\begin{equation}
g^*=\operatorname{clip}\!\left[\frac{\bar t}{\max(\bar B_0,10^{-3})},0.7,1.6\right].
\label{eq:s11}
\end{equation}

\begin{equation}
\begin{aligned}
a^*&=\operatorname{clip}(\bar t-g^*\odot\bar B_0,-0.08,0.08),\\
B^*&=\Pi\big(B_0\odot U(g^*)+U(a^*)\big).
\end{aligned}
\label{eq:s12}
\end{equation}

The teacher gain interval \([0.7,1.6]\) is wider than the student interval \([\exp(-0.28),\exp(0.28)]\). The teacher describes a coarse target adjustment, and the student fits it within its own bounded field family. Teacher-image loss is \(\operatorname{mean}(|B_1-B^*|)\). Gain and bias losses compare the coarse fields using smooth L1 with transition parameters 0.05 and 0.02, respectively. For a residual \(z\) and transition \(\beta\), smooth L1 equals \(z^2/(2\beta)\) when \(|z|<\beta\) and \(|z|-\beta/2\) otherwise. The \(10^{-3}\) denominator floor bounds ratios near a black base value, and the gain and bias clipping cap the teacher adjustment.

The smoothness term sums anisotropic total variation over \(g\) and \(a\): mean absolute neighboring differences along height plus the corresponding mean along width for each field. The dark-region weight is computed from the clipped base intensity as

\begin{equation}
d_0=\frac{\operatorname{ReLU}\big(1-\max\mathrm{RGB}(B_0)-0.28\big)}{0.72}.
\label{eq:s13}
\end{equation}

Dark-region L1 is the weighted absolute RGB error to \(y_p\) divided by the sum of the broadcast RGB weights, with a denominator floor of \(10^{-6}\). The direct image term is \(L_{\mathrm{direct}}=0.04\operatorname{MSE}(B_1,y_p)+0.03L_1(B_1,y_p)+0.10L_{\mathrm{dark}}\). For epoch \(k\), \(\rho(k)=0\) when \(k\leq24\) and \(\rho(k)=\min(1,(k-24)/40)\) otherwise. Together with the teacher-image, field, and total-variation terms, this gives main-text Eq. (9). The teacher image is computed analytically from training pairs. The 24-epoch teacher-only period trains the field representation, and the 40-epoch ramp introduces direct reference fitting. In Eq. \eqref{eq:s13}, the darkness threshold 0.28 makes the weight largest at zero base intensity and reduces it linearly to zero at \(\max\mathrm{RGB}(B_0)=0.72\).

\section{Training and Evaluation Details}\label{sec:training}

\subsection{Sequential optimization and checkpoint \mbox{selection}}

\begin{figure}[t]
\centering
\includegraphics[width=\columnwidth]{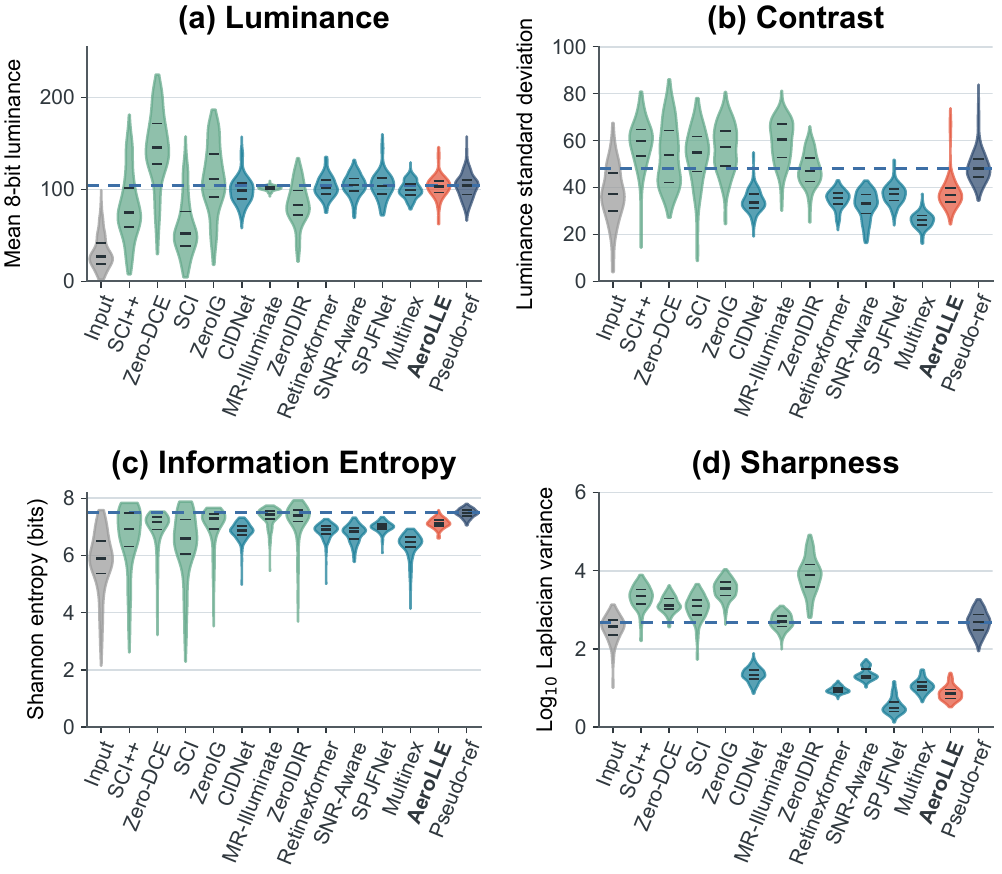}
\caption{Reference-free image statistics on the two AeroNight test sets. The plots show mean luminance, luminance contrast, information entropy, and Laplacian sharpness for the input, enhanced outputs, and screened pseudo-references. Entropy is reported for description and does not enter method ranking.}
\label{fig:s5}
\end{figure}

Stage 1 is initialized randomly and trained for 2,000 epochs on the 900 training pairs. It uses Adam with zero weight decay, batch size 8, \(384\times384\) paired crops, gradient clipping at 1, and a cosine learning-rate schedule initialized at \(3\times10^{-4}\). Random horizontal and vertical flips are synchronized across the input, reference, and scene thumbnail, and crop coordinates are updated consistently. The thumbnail covers the full source frame and supplies scene context for each local crop.

Internal selection is evaluated every 20 epochs through epoch 100, every 50 epochs thereafter, and at the final epoch. The selected Stage-1 checkpoint maximizes mean RGB uint8 PSNR over the 100 selection pairs. Ties retain the earlier checkpoint. This selected network is frozen before its training and selection outputs are exported.

Stage 2 is initialized randomly and optimized for 300 fixed epochs on whole images with batch size 2, without full-image flip or transpose augmentation. AdamW uses weight decay \(10^{-4}\) and gradient clipping at 1. A 10-epoch linear warm-up increases the learning-rate multiplier from 0.2 to 1.0 at a base rate of \(4\times10^{-5}\); cosine decay then reaches \(2\times10^{-6}\). The final epoch-300 checkpoint is used. Seed 3407 is used for the final training of both stages.

\subsection{Two-stage inference and image interface}

The selected Stage-1 model is fused and evaluated on full images. Clipped outputs are quantized to uint8 RGB and stored as PNG images. Stage 2 reloads them in \([0,1]\) for training and evaluation. Consistent with main-text Eq. (1), \(B_0=Q(\Pi(F_\theta(x)))\) with \(Q(v)=\lfloor255v+0.5\rfloor/255\). The floor expression specifies the implementation's rounding rule. This same quantized base supplies the calibration conditions, teacher construction, and affine application.

AeroLLE separates context-aware visibility recovery from spatial exposure--color calibration. Inference includes the fused local convolutions, thumbnail encoding and CSC, inter-stage quantization, coarse-field prediction, bilinear field upsampling, and full-resolution affine application. A single nighttime RGB image passes through these operations sequentially. The quantized interface can be reproduced by applying \(Q\) in memory. Parameter and FLOPs counts include both online stages.

The fused first-stage model contains 21,889 registered parameters, including its context encoder, latent filter, and CSC. The calibration head has 233,350 trainable parameters, giving 255,239 registered parameters in total. The first-stage count includes the 2,304 registered CSC transfer coefficients, whose diagonal entries are masked out.

\subsection{Fixed data roles and provenance}

The Base Enhancer is selected on the 100-pair internal model-selection split, and Test 1 and Test 2 are reserved for final evaluation. The comparison methods use the full 1,000-pair development pool, with checkpoints selected under their respective official configurations.

\subsection{Metric conventions and interpretation}

Paired evaluation computes PSNR per RGB image and averages the image scores, without border cropping or target-based brightness alignment. The evaluation tools record floating-tensor metrics and saved-uint8-image metrics separately. The AeroLLE Test 1 PSNR reported in Table 1 is evaluated from floating-point RGB outputs. SSIM for saved uint8 outputs uses a \(7\times7\) uniform window, RGB channel averaging, sample-covariance normalization, and data range 255; the valid interior of the SSIM window map is averaged. The table reports this measure as SSIM.

\subsubsection{Reference-free test-set descriptors}

Figure~\ref{fig:s5} summarizes four image-level descriptors on the two 100-image test sets at the original resolution. Saved outputs are converted to 8-bit grayscale with Pillow. Mean luminance and luminance standard deviation are the grayscale pixel mean and population standard deviation. Shannon entropy is \(-\sum_i p_i\log_2p_i\) over the 256-bin grayscale histogram, omitting zero-probability bins. The four-neighbor Laplacian is \(4Y\) minus its upper, lower, left, and right neighbors, using reflected boundaries. Sharpness is reported as \(\log_{10}(\max(\operatorname{Var}(\mathrm{Laplacian}),10^{-30}))\) with population variance. The violin plots report the first quartile, median, and third quartile of these image-level values. Entropy is a descriptive statistic and does not enter the method ranking.

\section{Additional Dataset Comparisons}\label{sec:datasets}

\subsection{Dataset scale and reference relationships}

Table~\ref{tab:s2} compares the scale, modality, and reference relationships of AeroNight and representative low-light datasets. Existing benchmarks cover paired RGB photographs, RAW short/long-exposure captures, multi-exposure stacks, detection images, and visible--infrared or event streams \cite{LOL,LOLv2,SID,SICE,ExDark}. AeroNight provides 1,200 RGB nighttime aerial pairs from multiple real UAV locations, with screened generated normal-light pseudo-references, 900 training pairs, 100 internal-selection pairs, and two held-out 100-pair test sets. The table highlights its focus on real nighttime aerial inputs and the distinct reference construction used in this benchmark.

\begin{table*}[t]
\centering
\includegraphics[width=\textwidth]{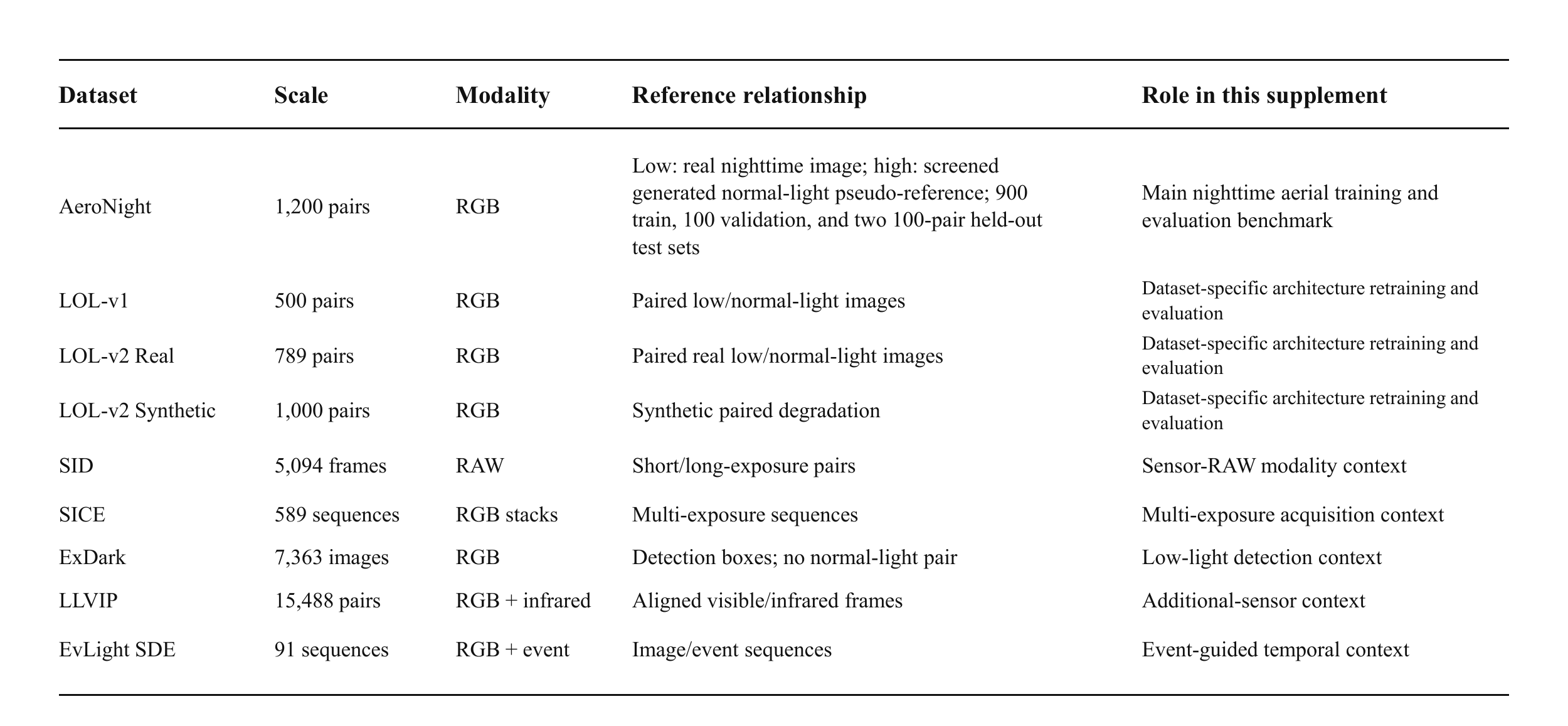}
\caption{Comparison of AeroNight with representative low-light and multimodal datasets. AeroNight contains 1,200 real nighttime UAV RGB inputs with screened generated normal-light pseudo-references, together with an internal-selection split and two held-out test sets.}
\label{tab:s2}
\end{table*}

\subsection{Aerial data without dataset-specific retraining}

Figure~\ref{fig:s9} compares enhancement results on U3D and VisDrone2019. Neither dataset provides a paired low-light training split for this task, so every method is applied using weights trained on AeroNight. This setting evaluates how the learned correction transfers to aerial scenes with different content and illumination patterns. On U3D, the supervised comparison methods generally increase the average brightness, but several outputs leave the vegetation and road surfaces around bright regions weakly exposed or spread the local illumination into neighboring areas. AeroLLE recovers the road, building boundaries, and vegetation surrounding the lamps while keeping the bright sources localized. It also maintains the gray pavement and green vegetation without the strong red or cyan shifts visible in some comparison outputs. On VisDrone2019, AeroLLE makes dark roads, roofs, and field boundaries easier to distinguish while preserving their spatial arrangement. The outputs retain the local bright regions and continue to show the darker roads and vegetation beside them, whereas some comparison methods either leave these regions underexposed or flatten the contrast around illuminated areas. These results show that the two-stage correction learned from AeroNight transfers to aerial scenes while preserving the scene-level and local exposure relationship.

\subsection{Non-aerial low-light datasets}

Figure~\ref{fig:s10} presents examples from LOL-v1, LOL-v2 Real, and LOL-v2 Synthetic. AeroLLE is trained on the training split of each dataset. On LOL-v1, the enhanced swimming pool, cupboard, and bowling alley show scene structures and colors close to their corresponding references. On LOL-v2 Real, the model reveals the air-conditioning unit, tables, and sports court in dark inputs and produces an overall appearance close to the normal-light images. On LOL-v2 Synthetic, details of buildings, vegetation, and the coastline become visible after enhancement, with exposure and color resembling the corresponding ground truth. These results illustrate the applicability of AeroLLE to both real and synthetic non-aerial low-light images.

\section{Additional Quantitative Analyses}\label{sec:quantitative}

\subsection{Random-seed evaluation}

A second training run with random seed 3411 achieves 17.5847 dB PSNR and 0.5168 SSIM on Test 1, and 17.7316 dB and 0.4969 on Test 2. Compared with the strongest competing results in Table 1 of the main paper, these values improve PSNR by 1.1753 dB and SSIM by 0.0014 on Test 1, and improve PSNR by 0.4765 dB and SSIM by 0.0012 on Test 2.

\subsection{Reference-free descriptors}

\begin{figure*}[!t]
\centering
\includegraphics[width=0.74\textwidth]{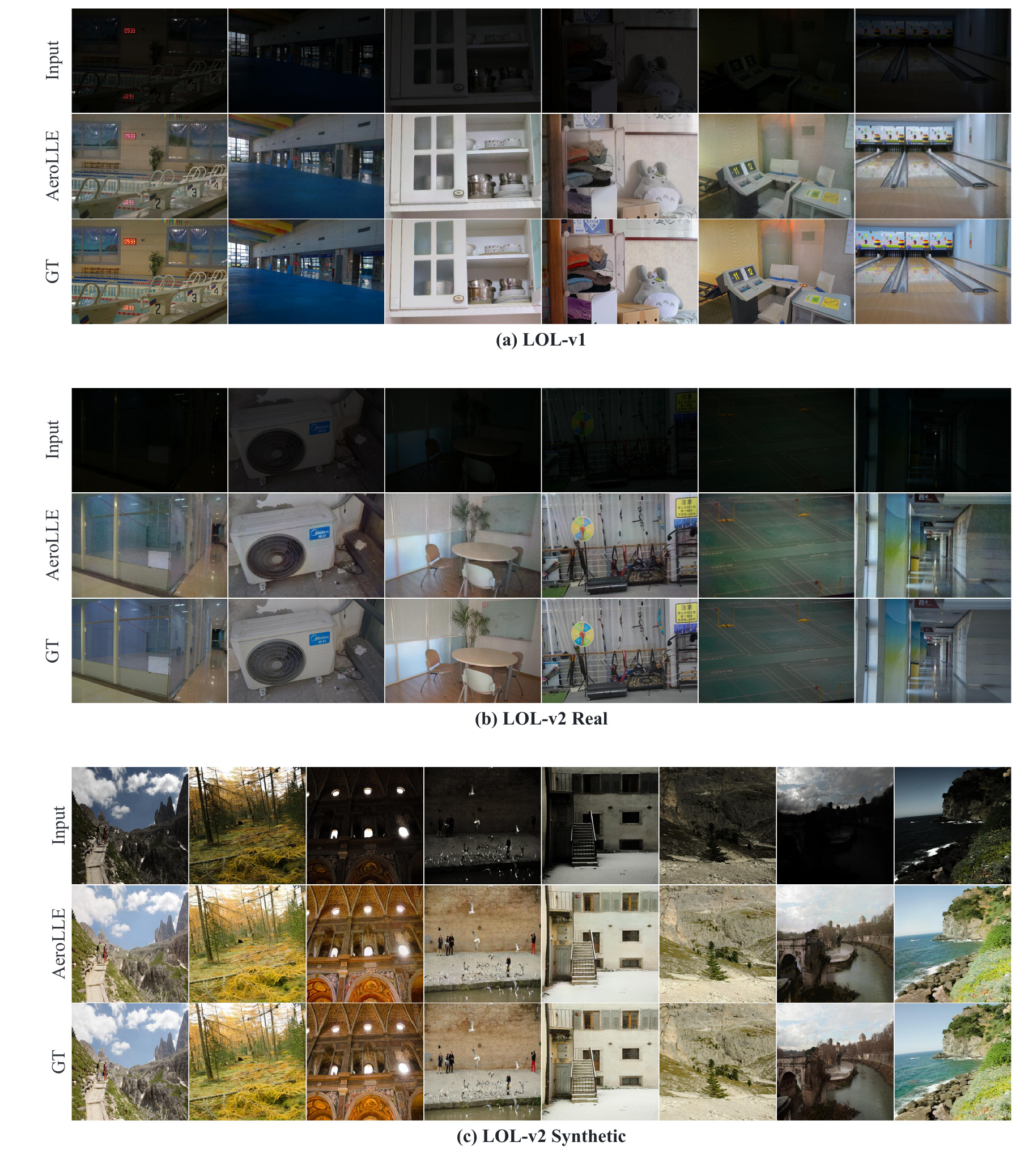}
\caption{Qualitative results on non-aerial low-light datasets. AeroLLE is trained on each dataset's own training split, and the corresponding reference image is shown as GT.}
\label{fig:s10}
\vspace{4pt}
\centering
\includegraphics[width=0.98\textwidth]{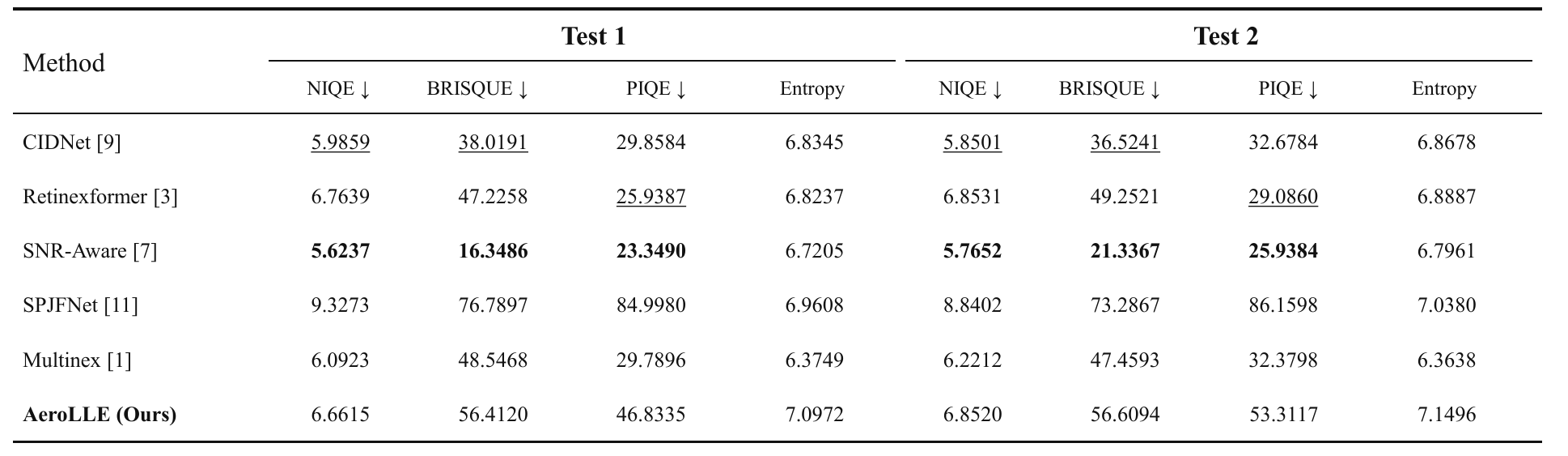}
\captionof{table}{Reference-free descriptors for supervised methods on the two AeroNight test sets. Lower is preferred for NIQE, BRISQUE, and PIQE. Entropy is reported as a descriptive statistic and does not enter method ranking.}
\label{tab:s4}
\end{figure*}

Table~\ref{tab:s4} reports NIQE, BRISQUE, PIQE, and entropy for the supervised methods and AeroLLE on both test sets. Entropy is descriptive and does not rank methods. The table shows that AeroLLE has the highest entropy among the listed methods on both test sets. The no-reference scores vary by descriptor and test set, so they are interpreted together with the paired PSNR/SSIM results and qualitative comparisons.

\section{Additional Qualitative Comparisons}\label{sec:qualitative}

\begin{figure*}[!t]
\centering
\includegraphics[width=0.94\textwidth]{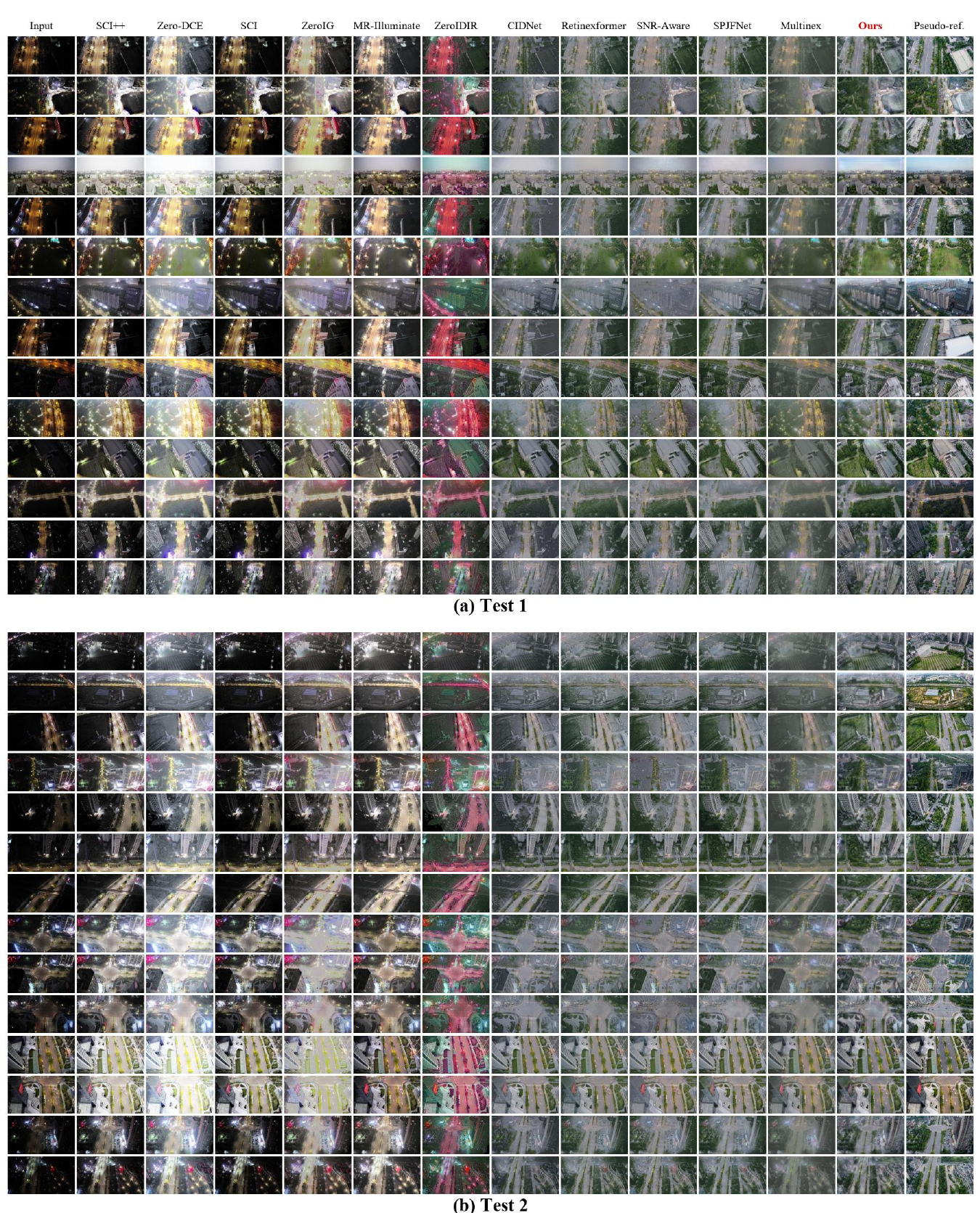}
\caption{Additional qualitative comparisons on AeroNight. Panel (a) shows 15 selected examples from Test 1, and panel (b) shows 15 selected examples from Test 2. Each row presents a nighttime input, the comparison methods, the final output of AeroLLE (Ours), and the generated pseudo-reference (Pseudo-ref.).}
\label{fig:s6}
\end{figure*}

Figure~\ref{fig:s6} presents 15 selected examples from each AeroNight test set. Panel (a) contains Test 1 examples from locations represented in development, and panel (b) contains Test 2 examples from locations outside the development scenes. The rows cover dark roads and vegetation, illuminated buildings, intersections, strong point lights, and large areas with uneven exposure. In the Test 1 examples, Zero-DCE often broadens bright bands around streetlights, while Multinex leaves nearby vegetation and road surfaces darker. AeroLLE raises the surrounding shadows while keeping the local intensity of lamps, so road boundaries, trees, and building edges remain visible. In the Test 2 examples, the method reveals field boundaries, roads, and building outlines while preserving the spatial layout and the relative brightness of already illuminated regions. Across both panels, the pseudo-reference guides appearance correction, while the captured nighttime input remains the source of scene structure.

{\small
\clearpage
\bibliographystyle{ieeenat_fullname}
\bibliography{references}
}
\end{document}